\pdfoutput=1
\documentclass[11pt]{article}

\usepackage[final]{acl}

\usepackage{times}
\usepackage{latexsym}

\usepackage[T1]{fontenc}
\usepackage[utf8]{inputenc}

\usepackage{microtype}

\usepackage{inconsolata}

\usepackage{graphicx}

\usepackage{threeparttable}
\usepackage{multirow}
\usepackage{multicol}
\usepackage{booktabs}
\usepackage{amsmath}
\usepackage{amssymb}
\usepackage{amsfonts}

\usepackage[utf8]{inputenc}
\usepackage{xcolor}
\usepackage{graphicx}
\usepackage{subcaption}

\usepackage{multicol}
\usepackage{multirow}
\usepackage{array}
\usepackage{xspace}
\usepackage{enumitem}
\setlist[itemize]{leftmargin=*}
\setlist[enumerate]{leftmargin=*}
\usepackage{amsmath}
\usepackage{amsfonts,amssymb}
\usepackage[font=small]{caption}
\usepackage{tablefootnote}
\usepackage{comment}
\usepackage{bm}
\usepackage{booktabs}
\usepackage{lineno}
\usepackage{array}
\usepackage{float}

\usepackage{placeins}

\usepackage[most]{tcolorbox}

\newcommand{\methodFont}{}
\newcommand{\ours}{\methodFont{Instructional Agents}\xspace}

\newcommand{\modelFont}{}
\newcommand{\gpto}{\modelFont{gpt-4o}\xspace}
\newcommand{\gptm}{\modelFont{gpt-4o-mini}\xspace}
\newcommand{\gptp}{\modelFont{o1-preview}\xspace}

\newcommand{\agentFont}{}
\newcommand{\faculty}{\agentFont{Teaching Faculty}\xspace}
\newcommand{\TA}{\modelFont{Teaching Assistant}\xspace}
\newcommand{\designer}{\modelFont{Instructional Designer}\xspace}
\newcommand{\coordinator}{\modelFont{Course Coordinator}\xspace}
\newcommand{\chair}{\modelFont{Program Chair}\xspace}

\usepackage{balance}
\usepackage{booktabs}

\title{\ours: Reducing Teaching Faculty Workload \\ through Multi-Agent Instructional Design}

\author{
  Huaiyuan Yao\textsuperscript{*},
  Wanpeng Xu\textsuperscript{*},
  Justin Turnau,
  Nadia Kellam,
  Hua Wei \\
  Arizona State University \\
  \texttt{\{huaiyuan, wanpeng.xu, jturnau, nadia.kellam, hua.wei\}@asu.edu} \\
}

\begin{document}
\maketitle

\def\thefootnote{*}\footnotetext{These authors contributed equally to this work}\def\thefootnote{\arabic{footnote}}
\begin{abstract}
Preparing high-quality instructional materials remains a labor-intensive process that often requires extensive coordination among teaching faculty, instructional designers, and teaching assistants. In this work, we present \ours, a multi-agent large language model framework designed to automate end-to-end course material generation, including syllabi creation, LaTeX-based slides, lecture scripts, and assessments. Unlike prior tools focused on isolated tasks, \ours simulates role-based collaboration to ensure pedagogical coherence. The system operates in four modes: Autonomous, Catalog-Guided, Feedback-Guided, and Full Co-Pilot mode, enabling flexible control over the degree of human involvement. We evaluate \ours across five university-level courses and show that it produces high-quality instructional materials that are reviewed and refined by teaching faculty prior to use, while significantly reducing the time required to prepare classroom-ready content. By supporting institutions with limited instructional design capacity, \ours provides a scalable and cost-effective framework to democratize access to high-quality education, particularly in underserved or resource-constrained settings. The project website, including source code, is available at \url{https://darl-genai.github.io/instructional_agents_homepage/}
\end{abstract}

\section{Introduction}

The preparation of instructional materials is a fundamental but labor-intensive aspect of education~\cite {merritt2016time, gavin2024teacher}. Instructors must design syllabi, create slides, and develop teaching notes, which often require coordination among faculty, instructional designers, and teaching assistants. Despite its pedagogical importance, the process is manual and time-consuming, limiting scalability. The absence of instructional design support exacerbates these challenges, often resulting in high preparation costs even for routine course development.

Recent advances in large language models (LLMs) have sparked growing interest in AI-assisted education~\cite{llmedu, gptedu}. While AI tools have addressed isolated tasks such as tutoring and grading~\cite{zhai2023chatgpt}, they lack end-to-end workflows for instructional design. As a result, instructors still invest substantial effort in producing coherent course materials, often resulting in fragmented alignment between objectives, assessments (e.g., quizzes, exams, and peer-reviewed assignments), and content~\cite{biggs1996enhancing, wang2013exploration, biggs2022teaching}.

To address these challenges, we introduce \ours, a multi-agent LLM framework for automated course material generation. Unlike single-model approaches, \ours simulates collaborative workflows among a comprehensive group of educational roles, including \faculty, \designer, \TA, \coordinator, and \chair. These agents interact guided by the instructional design framework, ADDIE~\cite{gagne2005principles, branch2009instructional}, ensuring alignment across learning objectives, assessments, and content. \ours also supports four modes: Autonomous, Catalog-Guided, Feedback-Guided, and Full Co-Pilot. These modes allow for a balance between automation and human involvement. 
By mimicking real-world instructional collaboration, the system aims to preserve instructional coherence while scaling the design process.

\begin{figure*}[t]
\centering
\includegraphics[width=\linewidth]{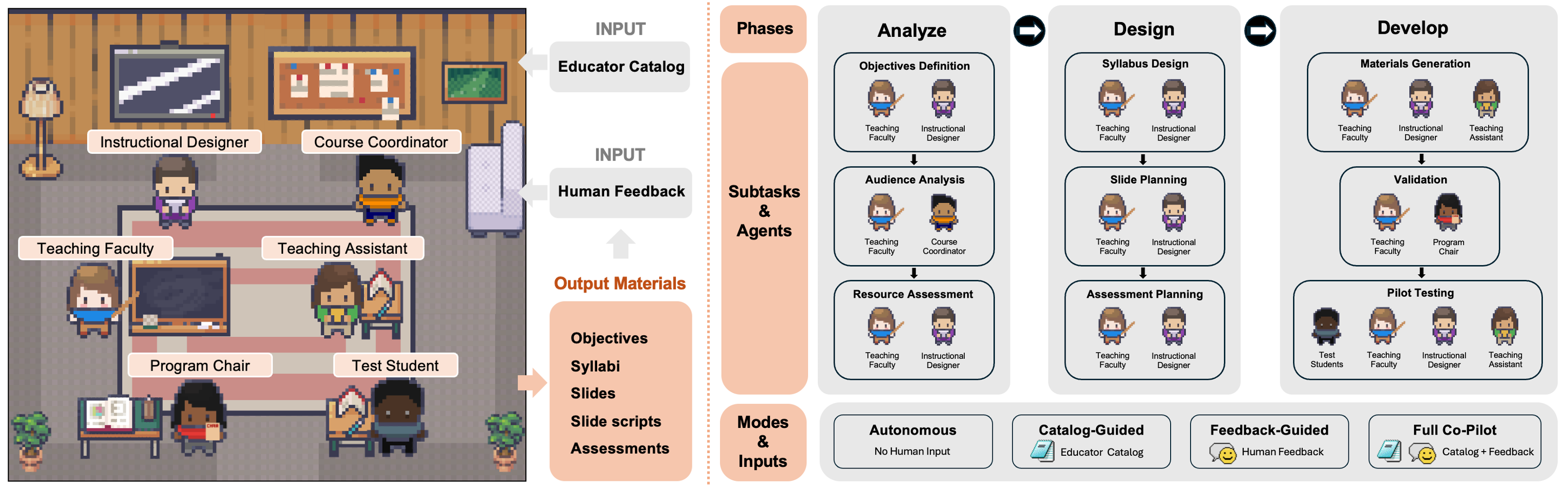}
\vspace{-5mm}
\caption{\textbf{Overview of \ours.}
(Left) Inputs and outputs in \ours. Educator input and human feedback guide the generation of key instructional materials, including learning objectives, syllabi, slides, slide scripts, and assessments.
(Right) Instructional Agents framework showing the overall workflow based on the first three phases of the ADDIE instructional design framework~\cite{gagne2005principles, branch2009instructional}: Analyze, Design, and Develop. Within each phase, multiple role-specialized agents (\faculty, \designer, \TA, \coordinator, and \chair) collaborate through structured prompt exchanges to complete subtasks and refine outputs in an iterative workflow. Appendix~\ref{sec:app:prompt} provides the specific prompts used for each type of agent. Each prompt includes a tailored background context and clearly defines the agent's goals, tasks, and responsibilities to ensure coherent and role-aligned response generation. Avatars are illustrative and designed for diversity without implying real demographic proportions or stereotypes.
}
\label{fig:instructional_agents_framework}
\vspace{-5mm}
\end{figure*}

This paper investigates whether multi-agent LLM systems can support instructional material generation in higher education. We evaluate how interaction modes impact output quality, efficiency, and scalability, with a focus on reducing faculty workload while preserving pedagogical rigor. Rather than student outcomes, we assess the revision effort required by teaching faculty.
In summary, our contributions are as follows:

\noindent$\bullet$~We present \ours, a multi-agent LLM framework for automated course material generation, structured around the ADDIE instructional design framework.

\noindent$\bullet$~We introduce role-based collaboration among educational agents with different levels of human involvement to ensure coherent, pedagogically aligned content. Specifically, \ours supports comprehensive roles in instructional design and operational modes to flexibly balance automation and human oversight.

\noindent$\bullet$~We evaluate \ours on five courses using both human and automated reviewers, showing that it reduces educator workload and preserves rigor and coherence, while also revealing trade-offs between automation, quality, and scalability.

\section{Background and Related Work}

\paragraph{LLM Agents and Role Collaboration}
Large Language Models have enabled the creation of multi-agent systems where model instances assume distinct roles to collaborate on complex tasks~\cite{yao2025comal, fuasagents}. These systems have been applied to domains such as scientific research~\cite{ma2024llm, schmidgall2501agent}, web automation~\cite{yang2024agentoccam}, and interactive behavior simulation~\cite{park2023generative, lilodriver}, demonstrating that structured agent interaction improves task consistency and division of labor~\cite{wang2025tdag, rasal2024navigating}. However, most applications optimize for factual correctness or task success, without addressing pedagogical alignment or coherence. The education domain presents distinct challenges, requiring collaboration across instructional roles and consistency among diverse outputs, which are not addressed by typical LLM agent pipelines~\cite{chu2025llm}.

\vspace{-2mm}
\paragraph{LLMs in Education}
In education, recent studies have focused on classroom simulation and task-specific automation. For example, LLM agents have been used to emulate teacher-student dialogues for training and research~\cite{zhang2024simulating, hao2025ai, hu2025exploring}. Others target automation of instructional tasks such as syllabus drafting, lesson planning, or content review~\cite{davis2023prompt, fan2024lessonplanner, roodsari2024instructional}. While these systems show promise, they often operate in isolation and lack integration into broader instructional pipelines. Current applications often fall short in ensuring educational rigor and alignment with pedagogical goals~\cite{kasneci2023chatgpt}. Recent work continues to critique AI applications in education for their limited pedagogical grounding and lack of integration with established instructional frameworks ~\cite{zawacki2024new}.

\vspace{-2mm}
\paragraph{Instructional Design and Automation}
Instructional design frameworks such as ADDIE~\cite{gagne2005principles, branch2009instructional} emphasize structured development across phases: Analyze, Design, Develop, Implement, and Evaluate. These frameworks offer a clear structure, but their real-world adoption is limited. Many instructors struggle to translate such models into practice due to time constraints and lack of support ~\cite{bennett2017process}. Others point to deeper institutional barriers, including insufficient incentives and tensions with professional identity ~\cite{brownell2012barriers}. While recent LLM-based tools have shown potential to automate isolated instructional tasks, they are typically single-pass and operate without reference to instructional design frameworks or cross-role coordination. Our approach embeds the ADDIE structure into a multi-agent LLM framework that simulates collaboration among instructional roles and supports pedagogically aligned material generation across the full course development pipeline.

\section{Method: System Design and Workflow}

To support collaborative instructional design, we propose \ours, a multi-agent LLM system that automates course content generation through role-specialized collaboration. The system simulates common educational roles involved in course development, including \faculty, \designer, \TA, \coordinator, and \chair. Among these roles, the Teaching Faculty agent serves as the primary authority and maintains continuous oversight throughout the entire workflow, while other agents provide complementary support for structure, implementation, validation, and feedback. Together, these agents operate within a structured workflow inspired by the ADDIE instructional design framework to produce a coherent and instructionally aligned course package, including learning objectives, syllabi, assessments, slide content, and slide scripts.

\subsection{Workflow Overview}

Figure~\ref{fig:instructional_agents_framework} illustrates the overall end-to-end workflow of the system. To clarify how the workflow operates in practice, we first summarize the full process before describing individual components in detail. The workflow consists of three sequential phases: \emph{Analyze}, \emph{Design}, and \emph{Develop}, corresponding to the first three stages of the ADDIE framework.

In the \emph{Analyze} phase, the Teaching Faculty agent leads the formulation of instructional objectives and instructional intent. The Course Coordinator agent supports this process by providing course-level context, constraints, and background information, such as student characteristics and resource limitations. Together, these activities result in an Instructional Foundation Report, which serves as a shared grounding artifact for all subsequent phases.

In the \emph{Design} phase, the Teaching Faculty agent continues to guide pedagogical decisions, ensuring alignment between objectives, content, and assessments. The Instructional Designer agent supports this phase by structuring the syllabi, organizing instructional flow, and refining assessment plans. The outputs of this phase include structured syllabi, key instructional points, and draft assessments that define the course's pedagogical plan.

In the \emph{Develop} phase, the Teaching Assistant agent generates concrete instructional materials, including slides, slide scripts, and finalized assessments, under the guidance of the Teaching Faculty agent. The Program Chair agent then reviews the generated materials from a broader program-level perspective to provide validation and suggestions, and the Test Student agent supplies simulated learner feedback to support iterative refinement. The final outputs of this phase are refined instructional objectives, syllabi, slides, slide scripts, and assessments, forming a cohesive and instructionally aligned course package.

While the ADDIE framework formally includes additional \emph{Implement} and \emph{Evaluate} phases, this work focuses on the first three phases due to practical and ethical considerations, including the need for human oversight before deploying AI-generated instructional materials to real students.

\subsection{Analyze Phase}

The Analyze phase focuses on understanding the instructional goals, learner profiles, and logistical constraints. It consists of three subtasks: 

\paragraph{Objectives Definition}
In this subtask, the \faculty and \designer collaborate to define competency-aligned course objectives. The \faculty agent initiates goal proposals based on domain knowledge, while the \designer ensures alignment with accreditation standards and instructional best practices. 

\paragraph{Audience Analysis}
In the \textit{Audience Analysis} subtask, the \faculty agent works with the \coordinator agent to build a learner profile by analyzing student backgrounds, prior knowledge, and challenges. This helps shape prerequisites and instructional strategies.

\paragraph{Resource Assessment}
In this subtask, the \faculty agent assesses teaching needs, and the \designer evaluates institutional constraints (e.g., platform compatibility). Together, they define feasible instructional strategies.

\subsection{Design Phase}
The Design phase organizes the course structure and assessment strategy. Agents collaborate to create syllabi with weekly topics, outline instructional methods, and align assessments with learning objectives. Feedback to support formative evaluation is also planned in this phase.

\paragraph{Syllabus Design}
As shown in Figure~\ref{fig:instructional_agents_framework},
in the \textit{Syllabus Design} subtask, the \faculty and \designer agents jointly develop course syllabi, including weekly topics, readings, and assignments. This subtask uses the previously defined objectives and learner profile to structure the course timeline. The output is syllabi that specify content coverage, assessment milestones, and delivery modes, which can be used for subsequent content development.

\begin{figure}[t]
    \centering
    \includegraphics[width=\linewidth]{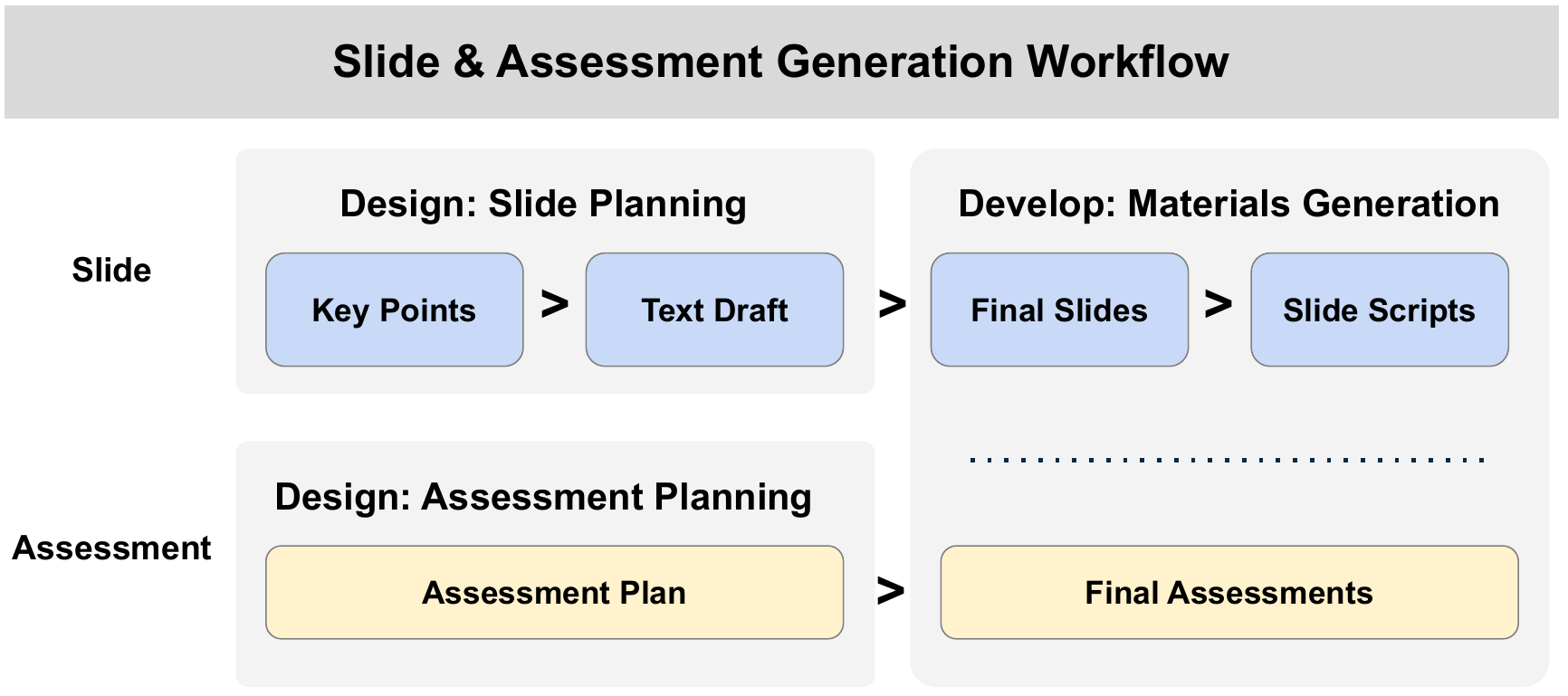}
\vspace{-6mm}
    \caption{Workflow of slide and assessment generation from key points and drafts to final slides, slide scripts, and assessments across the Design and Develop phases.}
    \label{fig:slide_assessment_workflow}
\vspace{-6mm}
\end{figure}

\paragraph{Slide Planning}
In this subtask, the \faculty and \designer agents co-develop the instructional flow for each weekly topic. The process begins with identifying key concepts and logical sequences based on the previously defined objectives and learner profile. The \faculty agent drafts initial slide content, including conceptual explanations, technical examples, and transitional narratives. The \designer agent then refines this content by structuring it for clarity, pedagogical flow, and visual coherence. The result is a slide content plan that serves as the foundation for material development in the subsequent phase. This process is visually summarized in the instructional workflow diagram shown in Figure~\ref{fig:slide_assessment_workflow}, which bridges the planning and generation stages across the Design and Develop phases.

\paragraph{Assessment Planning}
During the \textit{Assessment Planning} stage, the \faculty and \designer agents collaboratively define assessment strategies that align with course objectives. They design a multi-stage capstone project to replace traditional summative exams, incorporating deliverables such as a proposal, progress report, and final submission. Additionally, they establish formative feedback mechanisms, including peer review checkpoints, integrity guidelines, and grading rubrics. These assessments are integrated into the course timeline to ensure alignment with instructional goals and provide ongoing support for student learning. The output informs material generation and provides guidance for classroom use by \faculty.

\subsection{Develop Phase}

The Develop phase transforms the instructional plans from the Design phase into complete, classroom-ready materials. 
To ensure that the instructional outputs are pedagogically aligned, accurate, and usable, as shown in Figure~\ref{fig:instructional_agents_framework}, we design three interconnected stages: 

\paragraph{Materials Generation}
In this subtask, the \faculty, \designer, and \TA collaborate to generate all core instructional materials. For each chapter derived from the syllabi, the agents produce final LaTeX-based slides, slide scripts, and assessments.

The process begins by transforming slide planning outputs (i.e., key points and text drafts) into final slides and scripts. The \faculty agent expands the content with technical explanations and examples, the \designer structures the materials for pedagogical flow, and the \TA formats them into LaTeX documents. Similarly, assessment plans from the Design phase are converted into final assessments, including quizzes, milestones, and grading rubrics.

Once all content materials are finalized, the system performs a LaTeX compilation step to render the materials into publishable PDF packages. This integration process is handled by a dedicated \textit{LaTeX Compiler} module, which ensures consistent layout and formatting.

\paragraph{Validation}
The validation subtask involves expert review by the \faculty and \chair agents. All generated materials, including slides, slide scripts, and assessments, are reviewed for pedagogical alignment, content accuracy, and compliance with institutional expectations. The \chair agent provides suggestions and approval notes, which are incorporated by the \faculty or \designer agents before finalization. This ensures the materials meet program-level quality standards.
This step models how real instructors would revise and approve materials before using them in the classroom. No generated material is assumed ready for deployment without human oversight.

\paragraph{Pilot Testing}
To further evaluate usability, the system performs a pilot testing stage involving simulated student agents. These test agents engage with instructional materials under controlled scenarios. The \faculty, \designer, and \TA agents monitor the interactions and identify issues such as confusing phrasing, misaligned pacing, or missing prerequisite knowledge. Feedback collected during this stage informs final refinements before deployment.

\subsection{Modes of Operation}

\ours supports multiple modes of operation, each designed to accommodate different levels of human involvement and prior knowledge integration. The system can operate in four different modes, each with a different level of human input: Autonomous Mode (Auto), Catalog-Guided Mode (Cat), Feedback-Guided Mode (Feed), and Full Co-Pilot Mode (Pilot). These abbreviations are used in figures, tables, and other space-limited contexts for clarity.

\paragraph{Autonomous Mode}
In this mode, the system proceeds through all deliberations and content generation steps without human intervention beyond the initial course name or topic input. Each agent executes its role, moving from one subtask to the next upon completion. The agents autonomously generate learning objectives, syllabi, assessments, slides, and slide scripts. This mode is fully automated and suitable for baseline benchmarking or rapid prototyping of course content.

\paragraph{Catalog-Guided Mode}
Under this mode, the system incorporates pre-existing institutional or instructor-provided data as \texttt{Educator\_Catalog} to guide the deliberations. For example, predefined course structures, institutional policies, prior student feedback, and teaching constraints can be included in the \texttt{Educator\_Catalog} and passed to agents during initialization. These inputs inform agents' decisions, enabling the system to align outputs with existing curricula or departmental guidelines. This mode ensures continuity with institutional practices and reduces the risk of generating content that conflicts with prior standards. A sample catalog is provided in Appendix~\ref{sec:educator catalog sample}.

\paragraph{Feedback-Guided Mode}
This mode aims to enable retrospective correction and refinement of generated outputs. After a deliberation is completed, a human reviewer can inspect the results and provide targeted suggestions for improvement. The system supports rerunning individual deliberations with the new suggestions appended to the original context. This mode allows for iterative revision of specific materials, such as modifying assessment plans, without restarting the entire pipeline.

\paragraph{Full Co-Pilot Mode}
To simulate a collaborative workflow between the human teaching faculty and the agent system, in Full Co-Pilot Mode, the system pauses at the end of each subtask to solicit user feedback before proceeding. The user can approve the current outputs, request modifications, or provide guidance for the next steps. In addition to real-time feedback, this mode also incorporates structured preferences through the same \texttt{Educator\_Catalog} used in Catalog-Guided Mode. These catalog entries allow the system to maintain alignment with institutional policies and instructor intent across multiple subtasks, such as emphasizing specific topics in the syllabi, adjusting slide content focus, or altering assessment styles. By combining catalog initialization and human-in-the-loop feedback, Full Co-Pilot Mode closely mirrors real-world curriculum development, where iterative human review and prior knowledge are both integral to quality assurance.

\subsubsection{Summary of Modes}

These modes provide flexible control over the instructional design pipeline, ranging from fully autonomous execution to human-in-the-loop collaboration. Importantly, in all human-in-the-loop modes, Teaching Faculty retain control over final approval, ensuring that AI-generated content serves as a draft for human refinement. By enabling initialization from prior teaching artifacts, post-generation feedback integration, and interactive human collaboration, \ours supports a wide range of content development scenarios across instructional contexts.

\section{Experiments}

\begin{figure*}[t]
  \centering    \includegraphics[width=0.9\linewidth]{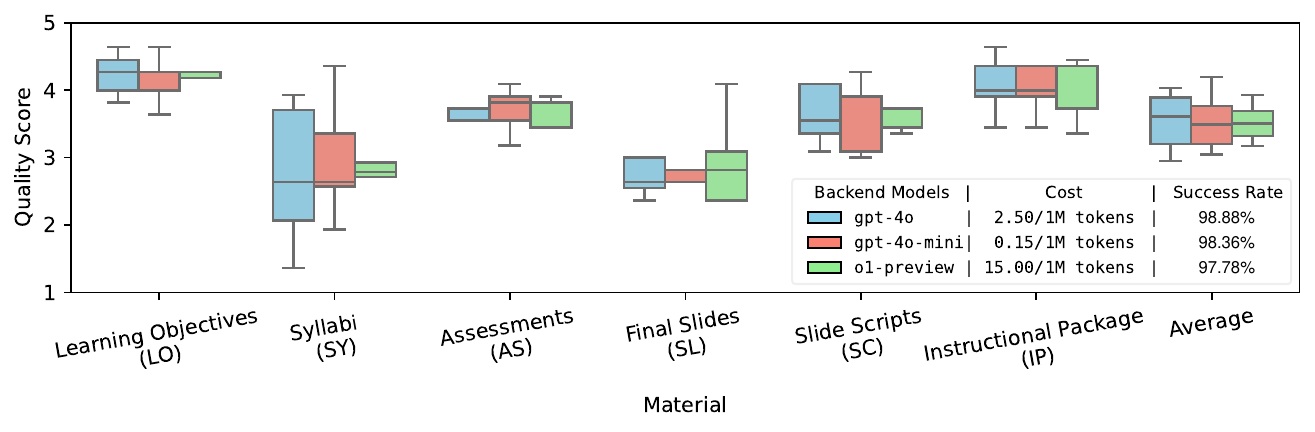}
  \vspace{-3mm}
  \caption{\textbf{(RQ1) Quality evaluation of generated instructional materials across different model backends with their costs and success rates.} 
This table reports the adapted QM-based rubric scores for course materials generated by \ours using three LLM backends: \gpto, \gptm, and \gptp. The evaluation is on six instructional outputs \ours generated: Learning Objectives (LO), Syllabi (SY), Assessments (AS), Final Slides (SL), Slide Scripts (SC), and the overall Instructional Package (IP). Scores are averaged over five human evaluators for each of the five courses. Each cell represents a score on a 1--5 Likert scale, where higher is better. \gptm achieves a performance level and success rate comparable to gpt-4o and o1-preview, while offering the lowest cost. Detailed numbers are provided in Appendix~\ref{sec:app:backend}.}
  \label{fig:model_comparision}
  \vspace{-5mm}
\end{figure*}

In this section, we present our experimental evaluation around three research questions:

\begin{itemize}[noitemsep]
    \item \textbf{RQ1:} How should we evaluate the quality of AI-generated instructional materials? How do human evaluations compare to LLM-based automated assessments?

    \item \textbf{RQ2:} How do different operational modes (Autonomous, Catalog-Guided, Feedback-Guided, Full Co-Pilot) affect instructional quality and instructor workload?

    \item \textbf{RQ3:} What are the runtime and cost trade-offs across different operational modes?
\end{itemize}

In this section, we additionally report ablations on different agent roles as well as success rate across different procedures. Further experimental results, including the influence of additional backend model evaluations (e.g., LLaMA, Qwen) and an ethics evaluation, are provided in Appendix~\ref{sec:additional_results}.

\subsection{Experimental Settings}
\paragraph{Model Backends}
We test the following model backends for content generation: \gpto~\cite{openai2024gpt4o}, \gptm~\cite{openai2024mini}, \gptp~\cite{openai2024preview}. 
To evaluate the framework, we apply \ours to five university-level courses that vary in structure and depth. The courses include Data Mining, Foundations of Machine Learning, Data Processing at Scale, Introduction to Artificial Intelligence, and Topics in Reinforcement Learning. Detailed hyperparameters are reported in Appendix~\ref{sec:app:hyperparameters}. We also test open-source models and report their results in Appendix~\ref{sec:app:open_source} and do not find that open-source models show superior performance. Therefore, in later parts of this paper, we primarily test using the above three GPT models.

\paragraph{Evaluation Criteria}
We adapt the Quality Matters (QM) Higher Education Rubric, Seventh Edition~\cite{qm2023rubric}, a widely used framework for quality assurance in online and hybrid course design, to evaluate instructional materials at the component level. While the original QM Rubric emphasizes holistic course structure, our version is guided by domain experts in instructional design and higher education and customizes selected QM dimensions to assess six key outputs generated by \ours: Learning Objectives (LO), Syllabi (SY), Assessments (AS), Final Slides (SL), Slide Scripts (SC), and the overall Instructional Package (IP). Each output is evaluated using a set of tailored metrics, such as clarity, alignment, and variety, which are designed to reflect its specific pedagogical role. Human evaluators rate each item on a 5-point Likert scale based on the revision effort required. The full scoring criteria, along with the mapping from Quality Matters (QM) dimensions to our adapted evaluation metrics, are provided in Appendix~\ref{appendix:rubric}.

\begin{figure*}[t]
  \centering    \includegraphics[width=0.99\linewidth]{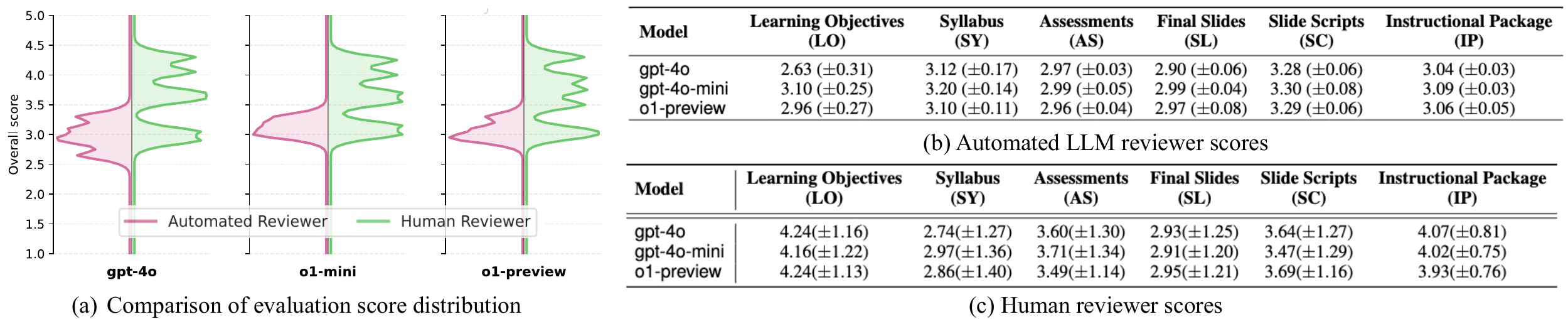}
  \vspace{-3mm}
  \caption{Comparison of evaluation scores (Human reviewer vs. Automated reviewer). 
  (a) The distribution of scores generated by human reviewer and automated reviewer. 
  (b) The scores of LLMs evaluating their own generated instructional materials. Each cell shows the mean (standard deviation) over five courses. Scores are on a 1--5 scale, where higher is better. 
  (c) The scores of human reviewers evaluating instructional materials generated different LLMs.
  Human reviewers tend to have more diverse evaluations while automated reviewers tend to give mediocre scores.
  }
  \label{fig:automated_and_human_reviewer}
  \vspace{-5mm}
\end{figure*}

\paragraph{Evaluator} We apply two kinds of evaluators: (1)~\textit{Human Reviewer}. For each course, we recruit five expert instructors to serve as human evaluators. These include faculty members and senior PhD students with prior teaching experience. Each evaluator rates the instructional package using six adapted criteria described above. (2)~\textit{Automated Reviewer}. In addition to human evaluation, we employ different LLMs (\gpto, \gptm, and \gptp) as automated reviewers to evaluate the generated materials using a rubric-based prompt aligned with the adapted QM-inspired metrics.% We test the automated evaluators with three different backends: \gpto, \gptm, and \gptp.

\subsection{Evaluating Instructional Quality: Human vs. LLM Reviewers (RQ1)}
We begin by addressing foundational questions about how to evaluate instructional content quality (RQ1), since this choice affects all subsequent experiments on comparisons of models and operational modes.
%
% \subsubsection{Analysis}
We analyze how different LLM backends influence the quality of generated materials and examine the alignment between automated LLM evaluations and human assessments. We report the following observations:

\noindent$\bullet$~\textbf{Influence of Backend Model:} Figure~\ref{fig:model_comparision} presents the evaluation results for six instructional materials and shows that all three backends produce high-quality content, with \gptm matching the performance of \gpto and \gptp while offering the lowest cost. A Friedman test confirmed no significant differences among the models (test statistic $Q = 0.473$, p-value $= 0.789$), supporting our conclusion~\cite{statistics}. Given this and the substantially lower computational cost and inference time of \gptm, we use it as the default backend model in the remainder of this paper. Detailed statistics are provided
 in Appendix~\ref{sec:app:backend}.

\noindent$\bullet$~\textbf{Human vs. LLM Evaluations:} Figure~\ref{fig:automated_and_human_reviewer} compares the distribution of overall evaluation scores assigned by automated LLM reviewers and human evaluators. LLMs consistently assign moderate and tightly clustered scores (typically between 2.9 and 3.1) while human evaluators produce a broader range of scores, demonstrating greater sensitivity to instructional effectiveness. This discrepancy highlights the limited capacity of LLM-based evaluators in distinguishing between higher and lower quality outputs~\cite{hong2026rulers}. Therefore, in the remainder of our experiments, we rely on human assessments as the primary reference for instructional quality. We also conducted the automated reviews across different LLMs, and the results are consistent with Figure~\ref{fig:automated_and_human_reviewer}. Their detailed results can be found in Appendix~\ref{sec:app:auto_reviews}.

\begin{table*}[ht]
\centering
\caption{
\textbf{(RQ2) Human evaluation on instructional materials across operational modes.} 
This table reports human ratings for course materials generated by \ours under four operational modes: Autonomous Mode (Auto), Catalog-Guided Mode (Cat), Feedback-Guided Mode (Feed), and Full Co-Pilot Mode (Pilot). Six key outputs generated by \ours are evaluated: Learning Objectives (LO), syllabi (SY), Assessments (AS), Final Slides (SL), Slide Scripts (SC), and the overall Instructional Package (IP). Each cell presents the mean rating averaged over five expert instructors per course. Scores are on a 1–5 scale, where the higher the better (Ratings reflect estimated instructor revision effort before classroom deployment). With greater human involvement, the material quality is better, and Full Co-Pilot mode consistently achieves the best performance.}
\label{tab:Human_evaluation_across_modes}
\vspace{-3mm}
\resizebox{\linewidth}{!}{
\begin{tabular}{c|cccc|cccc|cccc|cccc|cccc}
\toprule
\multirow{2}{*}{} 
& \multicolumn{4}{c|}{Course 1} 
& \multicolumn{4}{c|}{Course 2} 
& \multicolumn{4}{c|}{Course 3} 
& \multicolumn{4}{c|}{Course 4} 
& \multicolumn{4}{c}{Course 5} \\
\cmidrule{2-21}
& Auto & Cat & Feed                     & Co-Pilot       
& Auto & Cat & Feed                     & Co-Pilot           
& Auto & Cat & Feed                     & Co-Pilot  
& Auto & Cat & Feed                     & Co-Pilot  
& Auto & Cat & Feed                     & Co-Pilot                 \\
\midrule
LO & 3.73 & \underline{4.13} & 3.87 & \textbf{4.40} 
   & 3.87 & 4.07 & \underline{4.20} & \textbf{4.27} 
   & 3.42 & \textbf{4.17} & \underline{3.75} & \textbf{4.17} 
   & 3.17 & \textbf{3.75} & \underline{3.33} & \textbf{3.75} 
   & 3.58 & \underline{3.92} & 3.75 & \textbf{4.08} \\
SY & 3.10 & \underline{3.65} & 3.40 & \textbf{4.05} 
   & 2.90 & \underline{3.85} & 3.60 & \textbf{4.05} 
   & 2.81 & \underline{3.44} & 3.25 & \textbf{3.62} 
   & 2.94 & \underline{3.38} & 2.75 & \textbf{3.75} 
   & 2.94 & \underline{3.19} & \textbf{3.56} & \textbf{3.56} \\
AS & 3.10 & 3.45 & \underline{3.55} & \textbf{3.70} 
   & 2.95 & \underline{3.45} & \underline{3.45} & \textbf{3.70} 
   & 2.81 & \underline{3.31} & \underline{3.31} & \textbf{3.38} 
   & 2.38 & \textbf{3.31} & \underline{2.88} & \textbf{3.31} 
   & 2.31 & \underline{3.12} & 3.00 & \textbf{3.31} \\
SL & 2.87 & 3.20 & \underline{3.27} & \textbf{3.80} 
   & 3.00 & 3.13 & \underline{3.40} & \textbf{3.67} 
   & 3.00 & 3.17 & \underline{3.25} & \textbf{3.42} 
   & 2.58 & \textbf{3.42} & 3.25 & \underline{3.33} 
   & 2.50 & 3.08 & \textbf{3.42} & \underline{3.33} \\
SC & 3.20 & 3.47 & \underline{3.67} & \textbf{3.80} 
   & 3.67 & 3.73 & \textbf{4.13} & \underline{4.07} 
   & 3.17 & \underline{3.58} & 3.42 & \textbf{3.83} 
   & 3.08 & \underline{3.25} & \underline{3.25} & \textbf{3.33} 
   & 3.25 & 3.25 & \underline{3.42} & \textbf{3.50} \\
IP & 3.33 & \underline{3.87} & \underline{3.87} & \textbf{4.13} 
   & 3.07 & 3.40 & \textbf{3.87} & \underline{3.80} 
   & 2.75 & \textbf{3.50} & \underline{3.08} & \textbf{3.50} 
   & 2.25 & \underline{3.58} & 3.17 & \textbf{3.83} 
   & 2.50 & \underline{3.58} & 3.50 & \textbf{3.67} \\
\midrule
Avg & 3.22 & \underline{3.63} & 3.60 & \textbf{3.98} 
    & 3.24 & 3.61 & \underline{3.78} & \textbf{3.93} 
    & 2.99 & \underline{3.53} & 3.34 & \textbf{3.65} 
    & 2.73 & \underline{3.45} & 3.10 & \textbf{3.55} 
    & 2.85 & 3.36 & \underline{3.44} & \textbf{3.58} \\
\bottomrule
\end{tabular}
}
  \vspace{-5mm}
\end{table*}

\subsection{Impact of Operational Modes (RQ2)}

\begin{figure}[t]
    \centering
    \includegraphics[width=0.8\linewidth]{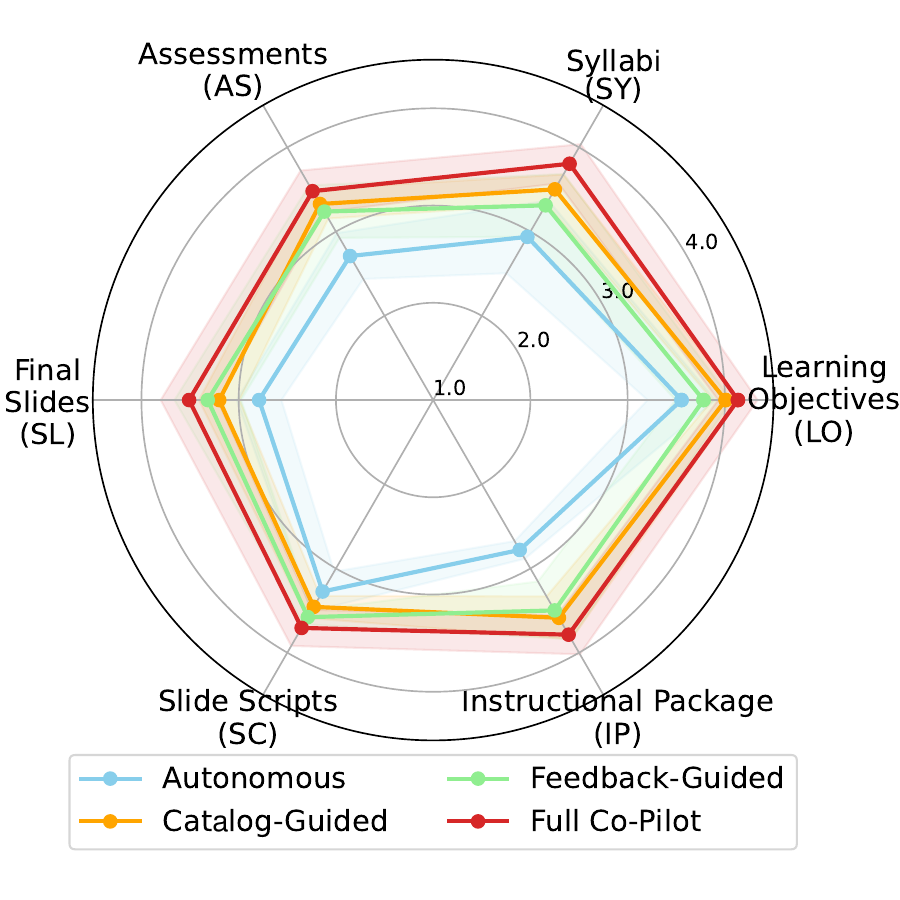}
    \vspace{-5mm}
    \caption{Radar chart analysis on the performance of generating materials at different modes. Each axis represents scores evaluated by human reviewers on one kind of material. Full Co-Pilot mode consistently performs the best.}
    \label{fig:radar}
  \vspace{-5mm}
\end{figure}

\begin{figure}[t]
    \centering
    \includegraphics[width=\linewidth]{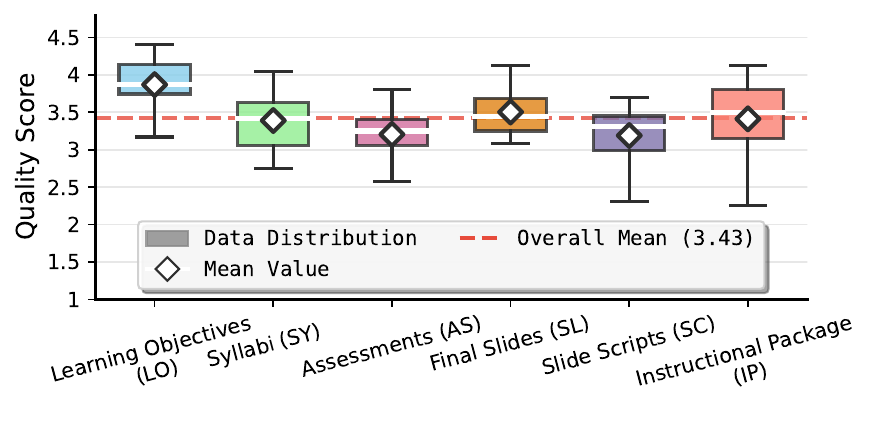}
    \vspace{-9mm}
        \caption{Rating distribution over instructional material types. }

        \label{fig:materials_boxplot}
  \vspace{-3mm}
\end{figure}

We examine how different operational modes in \ours influence instructional quality and human workload. The four operational modes differ in how they balance automation with human oversight. For each of the five courses, we generate instructional materials under all four modes and collect human ratings based on the adapted QM rubric. In addition to quantitative scores, we also collect open-ended qualitative feedback from evaluators, summarized in Appendix~\ref{sec:app:qualitative}. The results are shown in Table~\ref{tab:Human_evaluation_across_modes}, Figure~\ref{fig:radar}, and Figure~\ref{fig:materials_boxplot} . We present our key findings as follows:

\noindent$\bullet$~\textbf{Overall Comparison Across Modes:} In Table~\ref{tab:Human_evaluation_across_modes} and Figure~\ref{fig:radar}, Full Co-Pilot Mode consistently achieves the highest quality, improving scores by \textit{0.5 to 0.9 points} over Autonomous Mode, especially in Learning Objectives (LO), Slide Scripts (SC), and overall Instructional Packages (IP). Feedback-Guided Mode strikes a good balance between quality and efficiency, with stronger performance on content-rich components like Assessments (AS) and Slides (SL). In contrast, Catalog-Guided Mode outperforms Feedback-Guided Mode in components related to structure and administrative clarity, including Learning Objectives (LO) and syllabi (SY). This can be attributed to the use of pre-loaded templates and institutional guidelines, which support consistency but may limit depth and adaptability. These results highlight that human involvement improves quality, and each mode offers trade-offs between refinement and effort.

\noindent$\bullet$~\textbf{Material-level Trends:} As shown in Figure~\ref{fig:materials_boxplot}, all materials achieve average scores above 3.0, indicating generally acceptable quality across modes. Learning Objectives (LO) and Slides (SL)  receive the highest ratings on average, while Slide Scripts (SC) tend to score slightly lower. Notably, Slides (SL) also show lower variance across modes, suggesting that they are more robust to changes in workflow configuration.

\subsection{Runtime and Cost Analysis (RQ3)}
We evaluate the runtime efficiency and computational cost of \ours under different operational modes. The evaluation tracks four key metrics: Token Usage, Inference Time, Human Time, and Compute Cost. Human Time reflects instructor involvement for review or co-pilot interaction. All values are averaged across five courses using the \gptm backend. Table~\ref{tab:runtime_modes_gpt4o} summarizes the results. Autonomous Mode is the most efficient with lower-quality outputs. Catalog-Guided and Feedback-Guided Modes increase token usage slightly while requiring 10–30 minutes of teaching faculty effort. Full Co-Pilot Mode achieves the highest quality but requires the highest computational and human cost. These results highlight trade-offs between automation and quality: human-in-the-loop modes offer better instructional design outputs at the expense of time and effort. 

\begin{table}[htp]
\centering
\caption{
\textbf{(RQ3) Runtime and cost analysis across operational modes.} 
This table presents the runtime, token usage, human effort, and estimated compute cost for \ours using \gptm across four operational modes. The values are averaged over five courses. Inference time and token usage reflect resource consumption, while human time reflects instructor involvement for review or co-pilot interaction. Lower values are better for time, token usage, and cost.
}
\label{tab:runtime_modes_gpt4o}
\vspace{-2mm}
\resizebox{\linewidth}{!}{
\begin{tabular}{lcccc}
\toprule
\textbf{Metric} & Auto & Cat & Feedback                      & Co-Pilot  \\
\midrule
Token Usage (millions) & 1.46 & 2.05 & 1.93 & 2.42 \\
Inference Time (hrs) & 2.23 & 3.73 & 2.51 & 4.73 \\
Human Time (mins) & 0 & 10-15 & 20-30 & 30-45 \\
Compute Cost (USD) & 0.22 & 0.31 & 0.29 & 0.36 \\
\bottomrule
\end{tabular}
}
\vspace{-5mm}
\end{table}

\subsection{Ablation Studies on Different Agents}
To assess the contribution of each agent to the overall instructional design pipeline, we conduct detailed ablation studies by systematically removing individual roles from the multi-agent framework. As shown in Table \ref{tab:ablation_multi_metrics}, the single-agent baseline performs the worst overall (Avg = 2.33), highlighting the benefit of role specialization.
Removing the Teaching Faculty notably decreases syllabi (SY) and slide (SL) quality, since this agent provides domain knowledge and contextual grounding during the early stages of Analyze and Design.
Eliminating the Teaching Assistant results in a moderate quality drop in structural components such as slides and scripts, reflecting this agent’s critical role in formatting and LaTeX consistency.
The absence of the Instructional Designer causes a sharp decline in learning objectives (LO) and syllabi (SY) clarity, as this agent ensures pedagogical alignment and instructional structure across materials.

Together, these results demonstrate that role specialization is not merely an implementation choice, but a necessary design component for maintaining instructional coherence and quality across artifacts.

% certain outputs
\begin{table}[htbp]
\centering
\scriptsize   % 缩小字体，使表格更紧凑
\setlength{\tabcolsep}{3pt}  % 减少列间距
\caption{
\textbf{Ablation study on the role of agents.} 
}
\label{tab:ablation_multi_metrics}
\vspace{-2mm}
\begin{tabular}{lccccccc}
\toprule
\textbf{Method} & LO & SY & AS & SL & SC & IP & Avg \\
\midrule
Single Agent (GPT-4o-mini)   & 3.48 & 2.44 & 2.06 & 1.23 & 2.54 & 2.24 & 2.33  \\
w/o Teaching Faculty         & 3.53 & 2.12 & 2.83 & 1.67 & 3.83 & 2.50 & 2.75  \\
w/o Teaching Assistant       & 3.57 & 2.63 & 2.67 & 1.83 & 2.87 & 3.86 & 2.91  \\
w/o Instructional Designer   & 3.02 & 2.14 & 2.75 & 2.85 & 2.85 & 3.18 & 2.80 \\
Ours (Full, Auto)   & 3.55 & 2.93 & 2.71 & 2.79 & 3.27 & 2.78 & 3.01  \\
\textbf{Ours (Full, Co-Pilot)}   & \textbf{4.13} & \textbf{3.80} & \textbf{3.48} & \textbf{3.51} & \textbf{3.71} & \textbf{3.79} & \textbf{3.74} \\
\bottomrule
\end{tabular}
\vspace{-5mm}
\end{table}

\subsection{Success Rate Analysis}
\label{sec:app:success}

Detailed numbers on the distribution of success rates across different procedure can be found in Table~\ref{tab:app:success_rate}. Overall, gpt-4o demonstrates better reliability compared to gpt-4o-mini and o1-preview.

\begin{table}[H]
\centering
\resizebox{\linewidth}{!}{
\begin{tabular}{lcccccc}
\toprule
\textbf{Model} & 
\textbf{\begin{tabular}[c]{@{}c@{}}Learning Objectives \\ (LO)\end{tabular}}  &
\textbf{\begin{tabular}[c]{@{}c@{}}Syllabi \\ (SY)\end{tabular}} &
\textbf{\begin{tabular}[c]{@{}c@{}}Final Slides \\ (SL)\end{tabular}} & 
\textbf{\begin{tabular}[c]{@{}c@{}}Slide Scripts \\ (SC)\end{tabular}} & 
\textbf{\begin{tabular}[c]{@{}c@{}}Assessments \\ (AS)\end{tabular}} & 
\textbf{Avg} \\
\midrule
\gpto     & 100\% & 100\% & 94.4\% & 100\% & 100\% & 98.88\% \\
\gptm    & 100\% & 100\% & 91.8\% & 100\% & 100\% & 98.36\% \\
\gptp  & 100\% & 100\% & 88.9\% & 100\% & 100\% & 97.78\% \\
\bottomrule
\end{tabular}
}
\caption{Success rates (\%) of different models across various instructional design stages. 
Failures primarily stem from the generation of a small number of invalid or overly complex LaTeX codes, 
which lead to \texttt{pdflatex} compilation errors (although compilation often succeeds in Overleaf). }
\label{tab:app:success_rate}
\end{table}

\section{Discussion}
\label{sec:discussion}

In this section, we reflect the implications of \ours. For extended analysis, see Appendix~\ref{app:discussion}. \ours demonstrates that high-quality instructional materials can be generated with minimal human input and further enhanced through human-in-the-loop modes. Full Co-Pilot yields the best quality but requires more time and cost, while \gptm is the most efficient backend. These results support the system’s scalability in resource-constrained settings.

Ethical considerations are central to deployment. While LLMs may introduce bias, human review in Feedback-Guided and Full Co-Pilot modes helps ensure pedagogical soundness. \ours is designed to support, not replace, \faculty. Future work should address accessibility, originality verification, and inclusive content design. While it streamlines drafting, final content decisions rely on faculty judgment. Our evaluation quantifies the time \faculty are likely to save per component.
\section{Conclusion}

This paper presents \ours, a multi-agent LLM framework for automating the generation of instructional materials—syllabi, slides, slide scripts, and assessments—through simulated collaboration among educational roles. Evaluations across five courses show that while autonomous workflows reduce time and cost, incorporating human input, especially in Full Co-Pilot mode, improves quality and usability. Catalog-Guided and Feedback-Guided modes offer additional benefits in structural consistency and content depth.

\ours lowers the barrier to producing materials and enables scalable curriculum development in resource-constrained institutions. Reducing reliance on specialized support promotes broader access to instructional design. This is particularly impactful for community colleges, international programs, and underserved populations, where instructional capacity is limited. Through this work, we aim to support more inclusive, equitable, and globally accessible education systems.

% \clearpage
\section{Limitations}
\label{sec:limitations}

While our work demonstrates the potential of multi-agent LLM systems for automating instructional material generation, several limitations remain. First, the current framework primarily focuses on the \textit{Analyze}, \textit{Design}, and \textit{Develop} phases of the ADDIE model, without fully addressing \textit{Implementation} and \textit{Evaluation}, which require real-world classroom deployment and longitudinal assessment. Second, the current system has limited support for rich visual and interactive elements, which are important for modern pedagogy. Third, we do not treat bias analysis as a primary evaluation objective in this work. Although all generated materials are subject to faculty oversight, we only conduct an auxiliary bias evaluation using CEAT in Appendix~\ref{sec:ethics_evaluation}. Finally, the current system incorporates human feedback primarily through regeneration, rather than enabling fine-grained, targeted editing of specific content, which we leave for future work.

\section{Acknowledgments}
The work was partially supported by NSF award \#2442477. We thank Amazon Research Awards, Cisco Research Awards, Google, and OpenAI for providing us with API credits. The authors acknowledge Research Computing at Arizona State University for providing computing resources. The views and conclusions in this paper should not be interpreted as representing any funding agencies.

\bibliography{custom}

\appendix

\clearpage

% \appendix

\appendix

\setcounter{secnumdepth}{2}
\renewcommand{\thefigure}{A.\arabic{figure}}  
\renewcommand{\thetable}{A.\arabic{table}}
\setcounter{table}{0}  
\setcounter{figure}{0} 

\section{Hyperparameters}
\label{sec:app:hyperparameters}
Table~\ref{tab:agent_hyperparameters} summarizes the key hyperparameters adopted in our system of Instructional Agents.
The hyperparameters are grouped into two categories: (i) \textit{Foundation Model}, which controls the generative behavior of the underlying large language model (e.g., sampling temperature and token penalties), and (ii) \textit{Application}, which specifies task-level settings such as the number of deliberation rounds and the default length of generated slides. 
These parameters were chosen empirically based on preliminary experiments to balance generation diversity, reliability, and efficiency in instructional material creation.

\begin{table}[H]
  \centering
  \caption{Hyperparameters for Instructional Agents}
  \begin{tabular}{@{}llr@{}}
    \toprule
    \textbf{Category} & \textbf{Hyperparameter} & \textbf{Value} \\
    \midrule
    \multirow{4}{*}{\textbf{Foundation Model}} 
      & Temperature          & 1.0 \\
      & Top-$p$              & 1.0 \\
      & Presence penalty     & 0.0 \\
      & Frequency penalty    & 0.0 \\
    \midrule
    \multirow{2}{*}{\textbf{Application}}
      & Deliberation rounds  & 1 \\
      & Default slides length & 30 \\
    \bottomrule
  \end{tabular}
  \label{tab:agent_hyperparameters}
\end{table}

\renewcommand{\thetable}{B.\arabic{table}}
\setcounter{table}{0}

\section{Evaluation Criteria}
\label{appendix:rubric}

We adapt the Quality Matters (QM) Higher Education Rubric (Seventh Edition) to evaluate the quality and usability of instructional materials generated by \ours. Rather than adopting the rubric for formal accreditation, we extract selected QM dimensions and map them to output-specific evaluation criteria tailored to our system. Our evaluation focuses on the revision effort required to bring each output to a high-quality state.

Each instructional output is assessed by human raters using a 5-point Likert scale:

\begin{itemize}
    \item \textbf{5} – Minimal edits required; ready to use.
    \item \textbf{4} – Minor revisions needed; content is mostly solid.
    \item \textbf{3} – Moderate revisions needed in structure or clarity.
    \item \textbf{2} – Major restructuring or rewriting required.
    \item \textbf{1} – Complete redevelopment needed; not usable as-is.
\end{itemize}

Table~\ref{tab:qm-mapping-simple} shows the mapping between each evaluation metric and the original QM standard categories. The rubric is customized for each output type to reflect its instructional function and expected pedagogical alignment. 

The evaluation dimensions were designed in collaboration with domain experts in instructional design and higher education. The expert helped select and adapt relevant elements from the official QM rubric, ensuring that each metric is pedagogically grounded and practical for evaluating generated content. For each output type (e.g., syllabi, assessments), we identified key quality indicators that are both observable and actionable for human reviewers. The resulting rubric aims to balance instructional rigor with feasibility in large-scale human evaluation.

\begin{table*}[htbp]
\centering
\caption{\textbf{Evaluation Metrics: Description and Mapping to QM Standards.} 
This rubric adapts selected elements from the official Quality Matters (QM) Higher Education Rubric to fit the specific needs of evaluating AI-generated instructional materials. The adaptation was conducted in collaboration with an instructional design expert to ensure pedagogical validity. Each metric was mapped to appropriate QM categories and tailored to match the functional role of each output type.}
\label{tab:qm-mapping-simple}
\renewcommand{\arraystretch}{1.3}
\resizebox{!}{0.5\linewidth}{
\begin{tabular}{p{3.5cm} p{3.5cm} p{6cm} p{3.5cm}}
\toprule
\textbf{Key Outputs} & \textbf{Metric} & \textbf{Description} & \textbf{\shortstack[l]{Mapped to\\QM Standard}} \\
\midrule

\multirow{3}{=}{\textbf{Learning Objectives} \\ \textbf{(LO)}} 
& Clarity & Objectives are stated clearly and use learner-friendly language. & 2.3, 2.4 \\
& Measurability & Objectives include measurable verbs that allow for assessment. & 2.1, 2.2 \\
& Appropriateness & Objectives match the course level and are realistic for learners. & 2.2, 2.5 \\
\midrule

\multirow{4}{=}{\textbf{syllabi} \\ \textbf{(SY)}} 
& Structure & The syllabi clearly present the course purpose and structure. & 1.2, 1.1, 1.3 \\
& Coverage & The Syllabi include a complete and specific list of objectives. & 2.2 \\
& Accessibility & Technology, skills, and background requirements are clearly listed. & 1.5, 1.6, 1.7 \\
& Transparency of Policies & Academic policies are presented clearly and accessibly. & 1.4 \\
\midrule

\multirow{3}{=}{\textbf{Slides} \\ \textbf{(SL)}} 
& Alignment & Slides support the achievement of learning objectives. & 4.1 \\
& Appropriateness & Content matches learner needs and course level. & Extended from 4.2, 4.3 \\
& Accuracy & Content is factually correct and up to date. & 4.4 \\
\midrule

\multirow{3}{=}{\textbf{Slide Scripts} \\ \textbf{(SC)}} 
& Alignment & Script content matches and expands on the slides. & 4.1 \\
& Coherence & Scripts follow a logical flow and are easy to follow. & Extended from 4.2 \\
& Engagement & Scripts include examples or prompts to engage learners. & Extended from 4.2 \\
\midrule

\multirow{3}{=}{\textbf{Assessments} \\ \textbf{(AS)}} 
& Alignment & Assessments directly reflect the stated learning objectives. & 3.1 \\
& Clarity & Instructions, grading criteria, and expectations are clearly explained. & 3.2, 3.3, 3.6 \\
& Variety & Assessments use different formats to support diverse learners. & 3.4 \\
\midrule

\multirow{3}{=}{\textbf{Instructional Package} \\ \textbf{(IP)}} 
& Coherence & Materials work together as a unified, logically connected set. & General across 1–6 \\
& Alignment & All materials align with the course learning objectives. & 1.1, 2.1, 3.1, 4.1 \\
& Usability & Materials are easy to access, navigate, and use. & 6.1, 6.2, 6.3, 8.1, 8.2 \\
\bottomrule
\end{tabular}
}

\end{table*}

All scores reflect the estimated amount of instructor editing required before the materials are ready to teach.

\renewcommand{\thetable}{C.\arabic{table}}
\setcounter{table}{0}

\section{Educator Catalog Sample}
\label{sec:educator catalog sample}

\begin{table*}[htp]
\centering
\caption{\textbf{An Educator Catalog Sample (hypothetical)}}
\label{tab:educator_catalog_mapping}
\renewcommand{\arraystretch}{1.3}

\resizebox{!}{0.5\linewidth}{
\begin{tabular}{p{4cm} p{4cm} p{8cm}}
\toprule
\textbf{Category} & \textbf{Field} & \textbf{Sample Content} \\
\midrule
\multirow{3}{*}{Student Profile} 
& Student Background & Graduate-level students with diverse disciplinary and international backgrounds. \\
& Academic Performance & Generally strong academic readiness with varied prior exposure to ML and assessment styles. \\
& Learner Needs and Barriers & Familiar with Python; uneven experience with tools like Colab; some gaps in math background; benefits from simplified language and visuals. \\
\midrule
\multirow{3}{*}{Instructor Preferences} 
& Emphasis and Intent & Prioritizes real-world application of data science and analytics. \\
& Style Preferences & Structured instructional scripts, minimal slide clutter, and a professional but supportive tone. \\
& Assessment Focus & Project-driven assessment with open-ended tasks; traditional exams are not emphasized. \\
\midrule
\multirow{3}{*}{Course Structure} 
& Learning Outcomes & Ability to apply core ML methods (e.g., classification, clustering, dimensionality reduction) to real datasets with standard evaluation metrics. \\
& Duration & Semester-length offering. \\
& Weekly Outline & Example: Early weeks focus on data prep; middle on supervised and unsupervised learning; final weeks on applications and project presentations. \\
\midrule
\multirow{2}{*}{Assessment Design} 
& Format Preferences & Multi-stage project structure, minimal use of quizzes, open-ended problem-solving emphasized. \\
& Delivery Constraints & Submissions via learning management system in standard file formats (e.g., PDF, notebook). \\
\midrule
\multirow{4}{*}{Teaching Constraints} 
& Platform Policy & Hosted on institution-approved LMS; materials must meet accessibility and compliance standards. \\
& TA Support & Part-time teaching assistant available for grading and support. \\
& Delivery Context & In-person sessions with live walkthroughs and laptop-based activities. \\
& Max Slide Count & Approximately 50 instructional slides per course. \\
\midrule
\multirow{3}{*}{Institutional Requirements} 
& Program Outcomes & Aligns with learning goals related to modeling, evaluation, and practical data analysis. \\
& Academic Policies & Adheres to university-wide policies on integrity, accessibility, and digital content use. \\
& Department Syllabi & Required components include outcomes, grading policy, and institutional statements. \\
\midrule
Prior Feedback & Historical Evaluation Results & Feedback highlights preference for hands-on learning; pacing adjustments recommended in early weeks. \\
\bottomrule
\end{tabular}
}
\end{table*}

\noindent This educator catalog table organizes the instructional context into seven main categories: student profile, instructor preferences, course structure, assessment design, teaching constraints, institutional requirements, and prior feedback. Each field is mapped to representative content from a hypothetical instructor catalog. Details for each field and example values are provided in Table~\ref{tab:educator_catalog_mapping}. The table facilitates structured instructional design and can be used as a template for future course profiling and alignment with teaching goals.

% \medskip
% \noindent \textit{Note: This catalog sample is hypothetical and anonymized to preserve double-blind review integrity.}

\renewcommand{\thetable}{D.\arabic{table}}
\renewcommand{\thefigure}{D.\arabic{figure}}
\setcounter{table}{0}
\setcounter{figure}{0}

\section{Additional Results}
\label{sec:additional_results}

\begin{table*}[htp]
\centering
\caption{\textbf{Quality Evaluation of Generated Instructional Materials Across Model Backends.} 
This table reports the Quality Matters (QM) Rubric scores for course materials generated by \ours using three different LLM backends: \textbf{gpt-4o}, \textbf{gpt-4o-mini
}, and \textbf{gpt-o1-preview
}. The evaluation is based on an adapted rubric inspired by the Quality Matters (QM) framework, which we extend to assess six instructional outputs: Learning Objectives (LO), Syllabi (SY), Assessments (AS), Final Slides (SL), Slide Scripts (SC), and the overall Instructional Package (IP). Scores are averaged over five human evaluators for each of the five courses. Each cell represents a score on a 1--5 Likert scale, where \textbf{higher is better}. Detailed observations and analysis are provided in Section 4.1.
}
\label{tab:app:qm_backend_multicol}

\resizebox{\linewidth}{!}{
\begin{tabular}{lccccccc ccccccc ccccccc}
\toprule
\multirow{2}{*}{Course} & \multicolumn{7}{c}{\gpto} & \multicolumn{7}{c}{\gptm} & \multicolumn{7}{c}{\gptp} \\
\cmidrule(lr){2-8} \cmidrule(lr){9-15} \cmidrule(lr){16-22}
 & LO & SY & AS & SL & SC & IP & Avg & LO & SY & AS & SL & SC & IP & Avg & LO & SY & AS & SL & SC & IP & Avg\\
\midrule
A & 4.27 & 2.64 & 3.73 & 3.00 & 3.36 & 4.64 & 3.61 & 4.27 & 2.64 & 3.91 & 2.64 & 3.09 & 4.36 & 3.49 & 4.27 & 2.71 & 3.91 & 2.36 & 3.45 & 4.36 & 3.51\\
B & 4.45 & 3.71 & 4.09 & 2.64 & 4.09 & 4.36 & 3.89 & 4.27 & 3.36 & 4.09 & 2.64 & 3.91 & 4.36 & 3.77 & 4.27 & 2.79 & 3.82 & 3.09 & 3.73 & 4.45 & 3.69\\
C & 4.00 & 1.36 & 3.73 & 2.55 & 3.55 & 4.00 & 3.20 & 4.00 & 1.93 & 3.55 & 2.82 & 3.00 & 3.91 & 3.20 & 4.18 & 2.36 & 3.45 & 2.82 & 3.36 & 3.73 & 3.32\\
D & 3.82 & 2.07 & 2.91 & 2.36 & 3.09 & 3.45 & 2.95 & 3.64 & 2.57 & 3.18 & 2.36 & 3.09 & 3.45 & 3.05 & 3.82 & 2.93 & 2.82 & 2.36 & 3.73 & 3.36 & 3.17\\
E & 4.64 & 3.93 & 3.55 & 4.09 & 4.09 & 3.91 & 4.03 & 4.64 & 4.36 & 3.82 & 4.09 & 4.27 & 4.00 & 4.20 & 4.64 & 3.50 & 3.45 & 4.09 & 4.18 & 3.73 & 3.93\\
Avg & 4.24 & 2.74 & 3.60 & 2.93 & 3.64 & 4.07 & 3.54 & 4.16 & 2.97 & 3.71 & 2.91 & 3.47 & 4.02 & 3.54 & 4.24 & 2.86 & 3.49 & 2.95 & 3.69 & 3.93 & 3.52\\
\bottomrule
\end{tabular}}
 \end{table*}

\subsection{Influence of Backend Model}
\label{sec:app:backend}

Table~\ref{tab:app:qm_backend_multicol} (Quantitative Backend Comparison) shows the average quality scores of instructional outputs generated by \ours using three backend models: \gpto, \gptm, and \gptp. Across all six output types, \gpto and \gptm achieved comparable average scores (both 3.54), with \gptp slightly lower (3.52). Learning Objectives (LO) consistently scored highest across all backends, while syllabi (SY) and Final Slides (SL) showed greater variation and lower average scores. These trends suggest minor but observable differences in generation quality depending on the model backend.

\subsection{Comparison with Open-source Models}
\label{sec:app:open_source}
Table \ref{tab:ablation_baseline_multi_metrics} compares different backend models on six instructional design outputs using the adapted Quality Matters rubric. We include two recent open-source models (Qwen 2.5 72B and LLaMA 3.1 70B), Microsoft Co-Pilot, and GPT-4o-mini, which serves as our system backbone.

We do not adopt open-source models as primary backends because their performance on complex, multi-stage instructional design tasks remains consistently lower, despite recent progress on standard NLP benchmarks. Microsoft Co-Pilot performs competitively overall and slightly better on learning objectives (LO), but falls short on slide generation (SL = 2.40), reflecting its focus on short-form productivity rather than structured, long-context content creation.

% certain outputs
\begin{table}[htbp]
\centering
\scriptsize   % 缩小字体，使表格更紧凑
\setlength{\tabcolsep}{3pt}  % 减少列间距
\caption{
\textbf{Comparison with different backend models.} 
}
\label{tab:ablation_baseline_multi_metrics}
\vspace{-2mm}
\begin{tabular}{lccccccc}
\toprule
\textbf{Method} & LO & SY & AS & SL & SC & IP & Avg \\
\midrule
Qwen 2.5 72B         & 3.32 & 2.70 & 2.60 & 2.35 & 2.85 & 2.60 & 2.74 \\
LLaMA 3.1 70B       & 3.20 & 2.55 & 2.45 & 2.20 & 2.70 & 2.50 & 2.60 \\
Microsoft Co-Pilot & 3.60 & 2.95 & 2.75 & 2.40 & 3.10 & 2.85 & 2.94 \\
\textbf{GPT-4o-mini (Ours, Auto)} & 3.55 & 2.93 & 2.71 & 2.79 & 3.27 & 2.78 & 3.01 \\
\bottomrule
\end{tabular}
\vspace{-5mm}
\end{table}

\subsection{Automated Reviews Across Different LLMs}
\label{sec:app:auto_reviews}
\begin{figure*}[htbp]
\centering
\begin{subfigure}{1.0\linewidth}
    \centering
    \includegraphics[width=\linewidth]{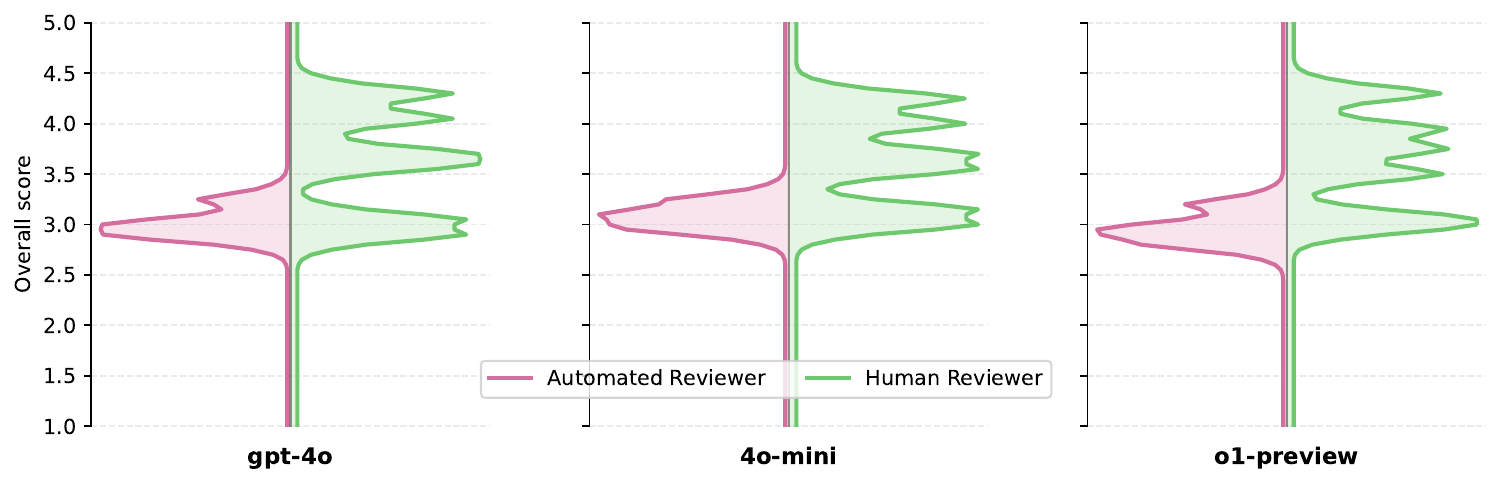}
    \caption{Automated LLM Reviewer: GPT-4o}
    \label{fig:subplot1}
\end{subfigure}
\begin{subfigure}{1.0\linewidth}
    \centering
    \includegraphics[width=\linewidth]{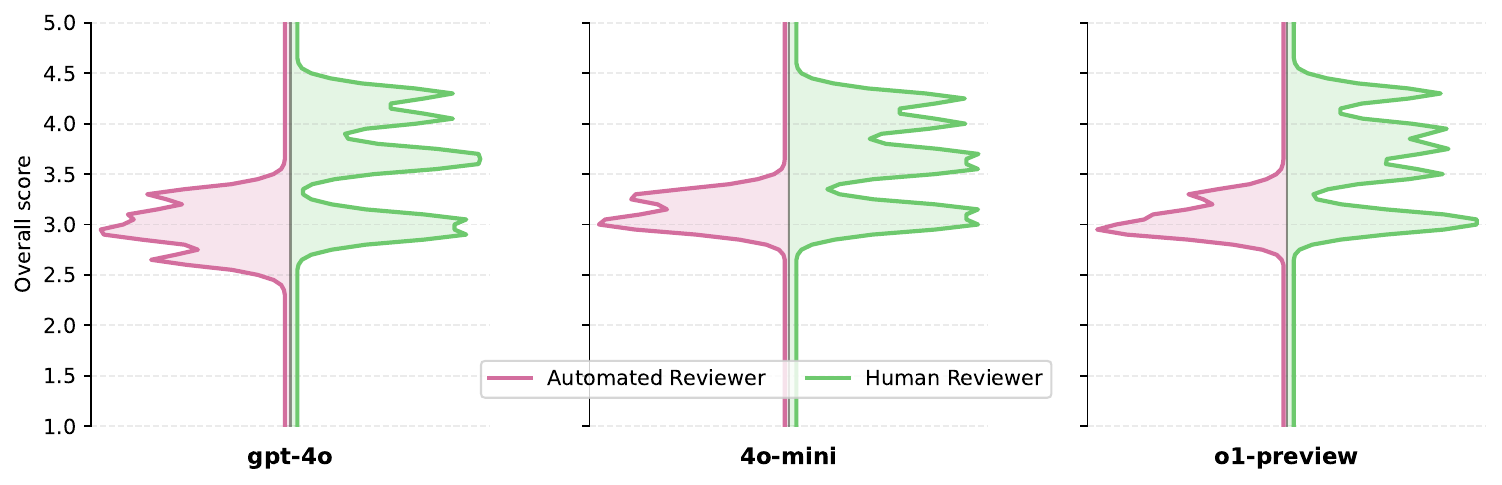}
    \caption{Automated LLM Reviewer: 4o-mini}
    \label{fig:subplot2}
\end{subfigure}
\begin{subfigure}{1.0\linewidth}
    \centering
    \includegraphics[width=\linewidth]{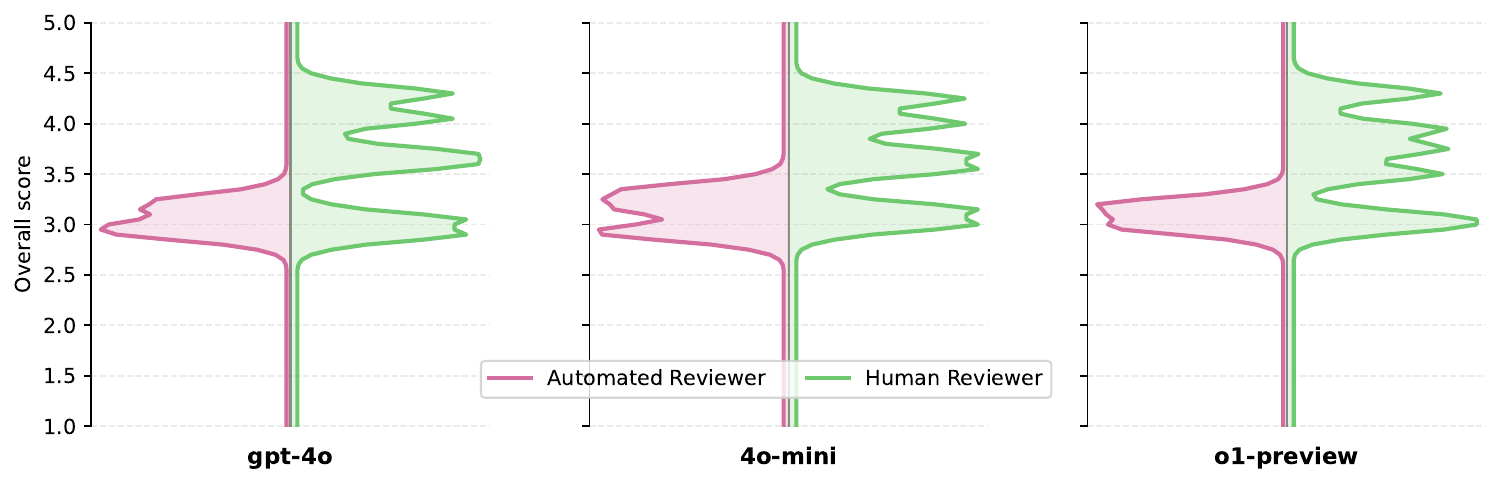}
    \caption{Automated LLM Reviewer: o1-preview}
    \label{fig:subplot3}
\end{subfigure}
\caption{Comparison of evaluation scores between human and automated reviewers across materials generated by different LLMs. In each plot, the horizontal axis shows the overall score distribution given by the reviewer (automated or human), and the vertical panels indicate which LLM generated the instructional materials (gpt-4o, 4o-mini, o1-preview). The three subfigures differ only in the automated reviewer model used. These plots highlight that automated reviewers produce tightly clustered scores, limiting their ability to distinguish between higher- and lower-quality outputs.}
\label{fig:three_score_distributions}
\end{figure*}

Figure~\ref{fig:three_score_distributions} compares evaluation score distributions between human reviewers and automated reviewers for materials generated by different backend models. Each subplot corresponds to a different LLM acting as the automated reviewer. Within each subplot, the vertical panels show materials generated by gpt-4o, gpt-4o-mini, and o1-preview, respectively. Across all reviewer backends, we observe that automated reviewers assign tightly clustered scores with limited variance, whereas human reviewers exhibit broader and more discriminative score distributions. This supports our earlier conclusion that LLM-based evaluators are less sensitive to content quality and therefore insufficient for high-stakes evaluation tasks.

\subsection{Quality of Materials and Mode-Based Analysis}
\label{sec:app:mode_analysis}

\begin{figure*}[htbp]
    \centering
    \includegraphics[width=0.7\linewidth]{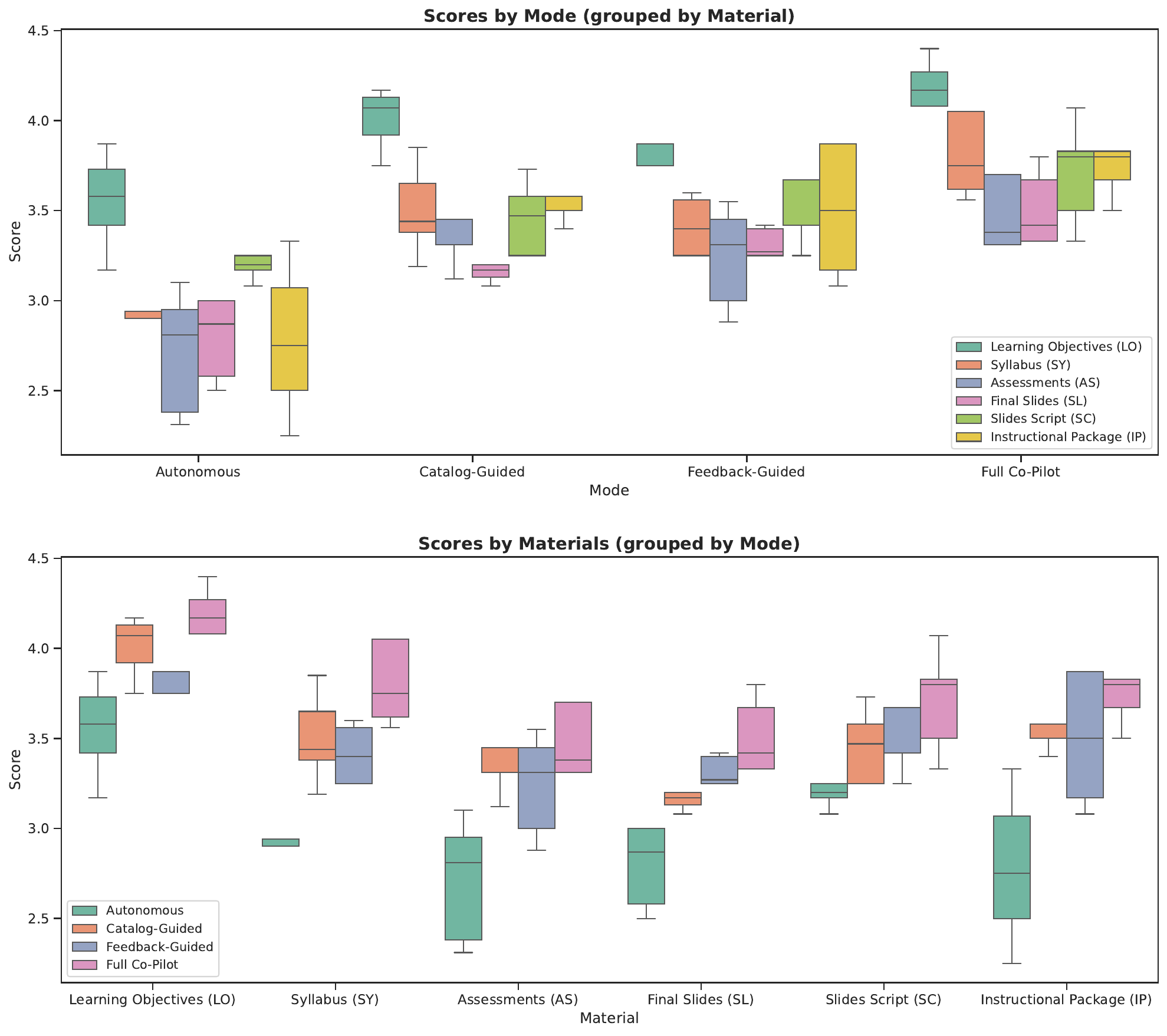}
    \caption{
Performance comparison across modes and materials.
\textbf{(a)} The Catalog-Guided and Feedback-Guided modes consistently outperform the Autonomous mode, while the Full Co-Pilot mode achieves the highest overall scores, demonstrating the effectiveness of our four-mode design.
\textbf{(b)} Higher scores are obtained for Learning Objectives (LO) and syllabi (SY), whereas performance on Slide Content (SL, SC) and Assessments (AS) is comparatively lower, indicating room for improvement in these materials.
}
    \label{fig:app:mode}
\end{figure*}

Figure~\ref{fig:app:mode} presents human evaluation results of instructional materials across different operation modes and material types. Subfigure~(a) shows that the Full Co-Pilot mode consistently achieves the highest overall quality, followed by Feedback-Guided and Catalog-Guided modes. The Autonomous mode yields the lowest median scores across most materials. Subfigure~(b) reveals that Learning Objectives (LO) and syllabi (SY) tend to receive higher ratings, while Final Slides (SL), Slide Scripts (SC), and Assessments (AS) exhibit greater variance and lower medians, indicating areas for improvement.

\subsection{Supplementary Bias Evaluation with CEAT}
\label{sec:ethics_evaluation}

We conducted a brief supplementary bias evaluation using the CEAT metric~\cite{peng2025auto}, a framework specifically designed for assessing potential bias in AI-generated educational content.
We applied this evaluation to materials produced by \texttt{gpt-4o-mini} across all four operational modes, with the goal of obtaining a general indication of whether any noticeable bias signals emerged.

To interpret the CEAT scores, we followed the effect size conventions described in the original CEAT work, where scores of 0.2, 0.5, and 0.8 are commonly interpreted as small, medium, and large effects, respectively~\cite{peng2025auto}.
These guidelines provide a straightforward basis for assessing the magnitude of any observed associations.

Across all tested categories, the CEAT scores remained below 0.2, which is the threshold generally considered indicative of a small effect.
Moreover, the mean p-values for all categories were substantially higher than the conventional significance level of 0.05, suggesting that the observed effects are not statistically significant.
Overall, we did not find evidence of notable or statistically significant bias in the generated materials.

\begin{table}[h]
\centering
\small
\begin{tabular}{lcc}
\toprule
Bias Category & CEAT Score & p-value \\
\midrule
National Bias & No bias detected & No bias detected \\
Racial Bias   & 0.0756 & 0.4416 \\
Gender Bias   & $-0.0612$ & 0.8921 \\
Other Bias    & $-0.1737$ & 0.8572 \\
\bottomrule
\end{tabular}
\caption{CEAT-based bias evaluation results across different bias categories.}
\label{tab:ceat_results}
\end{table}

\subsection{LaTeX Compilation Failures and Fixes}
\label{appendix:latex-errors}

While most instructional materials generated by \ours compiled without issues, some failures were caused by recurring LaTeX formatting errors. These were straightforward to identify and repair. The most common issues include:

\begin{itemize}
    \item \textbf{Missing \texttt{[fragile]} tag} when using verbatim-like content (e.g., code listings or unescaped symbols) inside frame environments.
    \item \textbf{Unescaped characters} such as \&, \%, and $\le$ in text or code.
    \item \textbf{Unicode symbols in math mode}, such as Greek letters (e.g., $\alpha$, $\beta$) or curved quotation marks, which must be replaced with corresponding LaTeX commands.
\end{itemize}

\begin{figure}[htbp]
\centering
\includegraphics[width=0.95\linewidth]{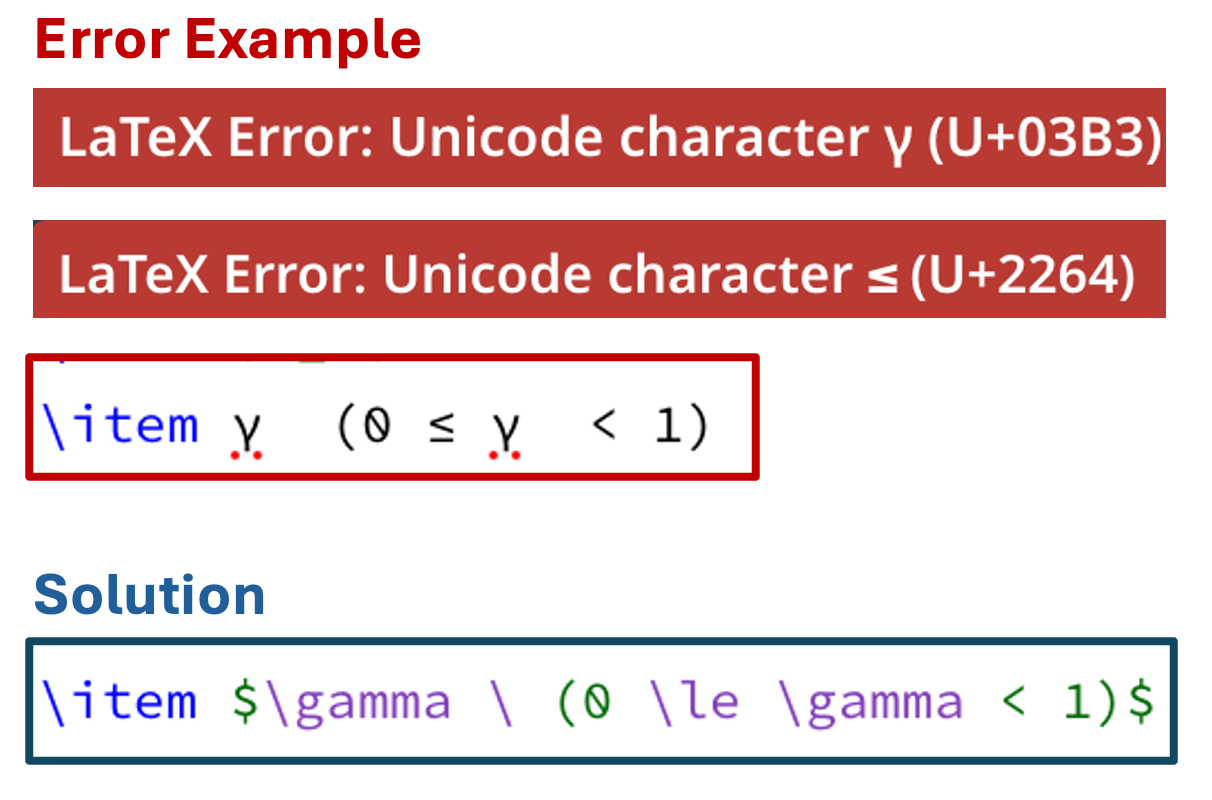}
\caption{\textbf{Example of Common LaTeX Errors and Fixes.} 
Top: raw output triggering compilation errors, such as unescaped Unicode characters 
(e.g., $\gamma$, $\le$). 
Bottom: corrected version using valid LaTeX syntax that compiles successfully.}
\label{fig:latex_errors}
\end{figure}

Figure~\ref{fig:latex_errors} presents an example showing both raw outputs that caused LaTeX compilation errors and their corresponding fixed versions. These issues were easily resolved without requiring expert LaTeX knowledge and did not affect the usability of outputs after minimal post-editing.

\subsection{Qualitative Feedback from Reviewers}
\label{sec:app:qualitative}
To complement our quantitative evaluation, we collected informal feedback from human reviewers regarding the perceived usefulness and limitations of the generated instructional materials. The feedback was gathered through optional open comment sections included in the rating forms and through brief follow-up conversations with several evaluators after the scoring sessions. In total, we engaged with five expert reviewers with prior teaching experience.

Overall, reviewers reported that the system meaningfully reduced the time and effort required to prepare core components such as syllabi and slides. One faculty member noted, \textit{``The generated materials gave me a strong foundation to build from and saved me substantial planning time."} Feedback-Guided and Full Co-Pilot modes were consistently highlighted as the most practical, particularly due to their support for iterative refinement. However, some instructors pointed out that the slide design templates were overly uniform, limiting visual variety and student engagement. These insights suggest that while \ours effectively accelerates content creation, user involvement and customizable visual outputs remain important for broader adoption in real classroom settings.

\section{Discussion}
\label{app:discussion}
In this section, we reflect on the strengths of \ours, summarize valuable insights from our experiments, discuss ethical considerations, and report observed failure cases and limitations.

\paragraph{Summary of Findings and System Strengths}
Our experiments show that \ours can generate coherent and pedagogically aligned instructional materials with minimal human input. Autonomous Mode achieves reasonable quality at minimal cost, while human-in-the-loop modes, especially Full Co-Pilot Mode, yield consistently higher scores, with improvements of 0.5–0.9 points on a 5-point scale. Among model backends, \gptm offers the best balance between quality and efficiency.

Across modes, we observe clear trade-offs: more human involvement improves quality but increases time and cost. LLM-based automatic evaluation shows limited reliability compared to human reviewers, who provide more consistent and sensitive assessments. Overall, \ours reduces faculty workload, supports content standardization, and enables scalable course development for resource-constrained educational settings.

\paragraph{Failure Cases and Limitations}

Despite the strengths of \ours, we observe several limitations and failure cases, particularly in the material generation subtask. A primary challenge arises from the LaTeX-based workflow: generated slides occasionally fail to compile due to syntax errors, template mismatches, or unsupported LaTeX packages. These issues were simple and repetitive, such as missing ``[fragile]''tags for code blocks, unescaped symbols like ``\&'', or the use of Unicode characters in math mode. All errors were quickly fixable by human reviewers without requiring LaTeX expertise, and detailed patterns and fixes are documented in Appendix~\ref{appendix:latex-errors}. These failures were most concentrated in the generation of Final Slides (SL), which depend heavily on LaTeX formatting. Appendix Table~\ref{tab:app:success_rate} reports success rates for SL rendering across different model backends.
While the textual content of slides is generally coherent, it often lacks visual or interactive elements. This reflects current LLM limitations in generating appropriate visual aids without explicit prompting and underscores the need for better integration with content rendering systems and future support for richer instructional design.

\paragraph{Ethical Concerns}
While \ours has the potential to streamline instructional material development, its deployment must be guided by ethical considerations. LLMs may reflect biases from training data, which could affect inclusivity or cultural sensitivity. To mitigate this, Feedback-Guided and Full Co-Pilot modes allow educators to review and revise content, preserving pedagogical alignment and minimizing unintended bias. \ours is intended to assist, not replace, instructors, by automating routine tasks while maintaining human oversight in course design. We also encourage instructors to verify originality and adhere to institutional policies on intellectual property. Lastly, although this work focuses on quality and coherence, future efforts should more directly address accessibility and equitable use across diverse learner populations. In addition to these mitigation mechanisms, we report a supplementary bias evaluation in Appendix~\ref{sec:ethics_evaluation} using the CEAT metric.

\balance
% \newpage
\section{Prompts}
\label{sec:app:prompt}
This appendix presents the prompt templates used for each role-specialized agent in the \ours framework.
Each prompt is designed to provide a tailored background context and to explicitly specify the agent’s objectives, responsibilities, and expected outputs.

% ================== Analyze Phase ==================
% Objectives Definition

\begin{figure*}[h]
\begin{tcolorbox}[breakable,enhanced,colback=blue!3!white,colframe=blue!70!black,
  title={Analyze --- Objectives Definition}]
\textbf{Interaction:} Teaching Faculty $\leftrightarrow$ Instructional Designer

\medskip
\textbf{Teaching Faculty}\par
{\small\ttfamily
You are a Teaching Faculty responsible for defining clear learning objectives based on accreditation standards, competency gaps, and institutional needs. Your goal is to draft a set of course objectives aligned with industry expectations and discuss with the department committee to refine them for curriculum integration.
}

\medskip
\textbf{Instructional Designer}\par
{\small\ttfamily
You are an Instructional Designer responsible for reviewing proposed learning objectives, assessing alignment with accreditation requirements, and suggesting modifications for consistency within the broader curriculum.
}
\end{tcolorbox}
\end{figure*}

% Audience Analysis
\begin{figure*}[h]
\begin{tcolorbox}[breakable,enhanced,colback=blue!3!white,colframe=blue!70!black,
  title={Analyze --- Audience Analysis}]
\textbf{Interaction:} Teaching Faculty $\leftrightarrow$ Course Coordinator

\medskip
\textbf{Teaching Faculty}\par
{\small\ttfamily
You are a Teaching Faculty responsible for identifying student learning needs based on prior knowledge, enrollment trends, and academic performance data. Your goal is to analyze gaps in student learning, assess common challenges, and discuss findings to ensure course design meets diverse student needs.
}

\medskip
\textbf{Course Coordinator}\par
{\small\ttfamily
You are a Department Admin responsible for providing institutional data on student demographics, enrollment trends, and past student feedback, then collaborating with professors to determine necessary course adjustments.
}
\end{tcolorbox}
\end{figure*}

% Resource Assessment

\begin{figure*}[h]
\begin{tcolorbox}[breakable,enhanced,colback=blue!3!white,colframe=blue!70!black,
  title={Analyze --- Resource Assessment}]
\textbf{Interaction:} Teaching Faculty $\leftrightarrow$ Instructional Designer

\medskip
\textbf{Teaching Faculty}\par
{\small\ttfamily
You are a Teaching Faculty responsible for determining the feasibility of courses based on faculty expertise, facility resources, and scheduling constraints. Your goal is to provide input on teaching requirements and ensure necessary instructional resources are available for effective course delivery.
}

\medskip
\textbf{Instructional Designer}\par
{\small\ttfamily
You are an Instructional Designer responsible for assessing whether current instructional technologies and platforms support proposed courses, identifying potential limitations, and collaborating to propose viable solutions.
}
\end{tcolorbox}
\end{figure*}

% ================== Design Phase ==================
% Syllabus Design

\begin{figure*}[t]
\begin{tcolorbox}[breakable,enhanced,colback=teal!3!white,colframe=teal!70!black,
  title={Design --- Syllabus Design}]
\textbf{Interaction:} Teaching Faculty $\leftrightarrow$ Instructional Designer

\medskip
\textbf{Teaching Faculty}\par
{\small\ttfamily

You are a Professor responsible for creating a structured syllabus that defines course content, pacing, and expected learning outcomes. Your goal is to draft a syllabus including weekly topics, learning objectives, required readings, and grading policies.
}

\medskip
\textbf{Instructional Designer}\par
{\small\ttfamily
You are a Department Committee Member responsible for reviewing syllabus drafts, assessing alignment with institutional policies and accreditation requirements, and providing recommendations for improvement.
}
\end{tcolorbox}
\end{figure*}

% Slide Planning (three agents)

\begin{figure*}[t]
\begin{tcolorbox}[breakable,enhanced,colback=teal!3!white,colframe=teal!70!black,
  title={Design --- Slide Planning}]
\textbf{Interaction:} Teaching Faculty $\leftrightarrow$ Instructional Designer $\leftrightarrow$ Teaching Assistant

\medskip
\textbf{Teaching Faculty}\par
{\small\ttfamily
You are a Teaching Faculty responsible for creating detailed educational content for slides. Your goal is to explain concepts clearly, provide examples, and make complex topics accessible to students.
}

\medskip
\textbf{Instructional Designer}\par
{\small\ttfamily
You are an Instructional Designer responsible for organizing course content into a logical slide structure. Your goal is to create an outline that covers all key topics with appropriate depth and flow.
}

\medskip
\textbf{Teaching Assistant}\par
{\small\ttfamily
You are a Teaching Assistant responsible for creating LaTeX slides and detailed speaker notes. Your goal is to create well-formatted slides and comprehensive speaking notes that explain all key points clearly.
}
\end{tcolorbox}
\end{figure*}

% Assessment Planning

\begin{figure*}[h]
\begin{tcolorbox}[breakable,enhanced,colback=teal!3!white,colframe=teal!70!black,
  title={Design --- Assessment Planning}]
\textbf{Interaction:} Teaching Faculty $\leftrightarrow$ Instructional Designer

\medskip
\textbf{Teaching Faculty}\par
{\small\ttfamily
You are a Professor responsible for designing a course's assessment and evaluation strategy. Your task is to define project-based, milestone-driven, and real-world-relevant assessments, including formats, timing, grading rubrics, and submission logistics. Avoid traditional exam-heavy approaches.
}

\medskip
\textbf{Instructional Designer}\par
{\small\ttfamily
You are a Department Committee Member responsible for evaluating assessment plans to ensure they align with institutional policies, learning outcomes, and best practices in competency-based education. Provide constructive feedback on assessment design, balance, and fairness.
}
\end{tcolorbox}
\end{figure*}

% ================== Develop Phase ==================

% ---------- Materials Generation : SLIDES ----------

\begin{figure*}[h]
\begin{tcolorbox}[breakable,enhanced,
  colback=purple!3!white,colframe=purple!70!black,
  title={Develop --- Materials Generation: Slides}]
\textbf{Interaction:} Instructional Designer $\leftrightarrow$ Teaching Faculty $\leftrightarrow$ Teaching Assistant

\medskip
\textbf{Instructional Designer}\par
{\small\ttfamily
Based on the chapter title and description, produce a detailed slides outline in valid JSON. 
Cover all key aspects with about \textit{N} slides; ensure structure is comprehensive and easy to follow; 
use simple, common LaTeX grammar so later compilation is robust.
}

\medskip
\textbf{Teaching Faculty}\par
{\small\ttfamily
For each slide (with context from adjacent slides), write clear, student-oriented educational content: 
explanations, examples/illustrations, key points, and any formulas/code/diagrams as text descriptions. 
Keep the content concise enough to fit a single PPT slide while aligning with chapter learning objectives.
}

\medskip
\textbf{Teaching Assistant}\par
{\small\ttfamily
Transform the outline and content into compilable Beamer LaTeX. 
Create frame placeholders per slide and, when needed, multiple frames for one slide; 
summarize the content into a brief lead-in, use lists/blocks/code/math environments appropriately, 
avoid non-ASCII symbols, and escape special characters. 
Output must be directly compilable LaTeX.
}
\end{tcolorbox}
\end{figure*}

% ---------- Materials Generation : SCRIPT (updated to three agents) ----------

\begin{figure*}[t]
\begin{tcolorbox}[breakable,enhanced,
  colback=purple!3!white,colframe=purple!70!black,
  title={Develop --- Materials Generation: Script}]
\textbf{Interaction:} Teaching Faculty $\leftrightarrow$ Instructional Designer $\leftrightarrow$ Teaching Assistant

\medskip
\textbf{Teaching Faculty}\par
{\small\ttfamily
Provide the technical and domain-accurate talking points for each slide; highlight the intended learning objectives, 
key takeaways, examples, analogies, and must-mention caveats; suggest transitions to previous/next slides; 
review draft scripts for accuracy and depth.}

\medskip
\textbf{Instructional Designer}\par
{\small\ttfamily
Shape the narrative and pacing for cognitive load management; ensure the script aligns with the slide outline, 
uses accessible language, and embeds engagement prompts (checks for understanding, rhetorical questions, brief activities); 
enforce consistency across multi-frame slides and coherence between sections.}

\medskip
\textbf{Teaching Assistant}\par
{\small\ttfamily
Synthesize inputs into a presenter-ready speaking script for each slide; 
reference the final LaTeX frames to insert clear cues for frame advances and timing; 
deliver a clean JSON/markdown script artifact per slide, integrate feedback from Faculty and Designer, 
and maintain smooth transitions and self-contained explanations for others to present effectively.}
\end{tcolorbox}
\end{figure*}

% ---------- Materials Generation : ASSESSMENTS (updated to three agents) ----------

\begin{figure*}[t]
\begin{tcolorbox}[breakable,enhanced,
  colback=purple!3!white,colframe=purple!70!black,
  title={Develop --- Materials Generation: Assessments}]
\textbf{Interaction:} Teaching Faculty $\leftrightarrow$ Instructional Designer $\leftrightarrow$ Teaching Assistant

\medskip
\textbf{Teaching Faculty}\par
{\small\ttfamily
Define what knowledge and skills each slide should assess; propose concept targets, real-world tasks, 
answer rationales, and expected solution sketches; calibrate difficulty and ensure content validity.}

\medskip
\textbf{Instructional Designer}\par
{\small\ttfamily
Map each item to learning objectives and Bloom levels; ensure variety and fairness (MCQ, short answer, practical tasks, 
discussion prompts), accessibility and bias checks, and alignment with program/policy; 
suggest rubric criteria and milestone integration.}

\medskip
\textbf{Teaching Assistant}\par
{\small\ttfamily
Produce the assessment artifacts per slide in valid JSON/markdown: 
3--5 MCQs with four options, correct answers and explanations; practical activities/exercises; 
learning objectives and discussion questions; apply formatting constraints and integrate Faculty/Designer feedback; 
compile a coherent assessment pack ready for delivery and LMS ingestion.}
\end{tcolorbox}
\end{figure*}

% ---------- Validation ----------
\begin{figure*}[!t]
\begin{tcolorbox}[enhanced,
  colback=green!3!white,colframe=green!60!black,
  title={Develop --- Validation}]
\textbf{Interaction:} Program Chair $\leftrightarrow$ Test Student

\medskip
\textbf{Program Chair}\par
{\small\ttfamily
Evaluate course materials for academic rigor and standards, alignment with program requirements, 
quality of instructional design, assessment validity/reliability, and overall coherence/structure. 
Provide detailed evaluation and constructive feedback.
}

\medskip
\textbf{Test Student}\par
{\small\ttfamily
Evaluate materials for clarity and understandability, engagement and motivation, learning support and guidance, 
practical applicability, and accessibility/user experience. 
Provide feedback from a student’s perspective.
}

\end{tcolorbox}
\end{figure*}

\end{document}